\documentclass{article} 
\usepackage{iclr2027_conference,times}

\usepackage{amsmath,amsfonts,bm}

\def\eqref#1{equation~\ref{#1}}

\def\1{\bm{1}}

\DeclareMathAlphabet{\mathsfit}{\encodingdefault}{\sfdefault}{m}{sl}
\SetMathAlphabet{\mathsfit}{bold}{\encodingdefault}{\sfdefault}{bx}{n}

\usepackage[utf8]{inputenc}
\usepackage[T1]{fontenc}
\usepackage{xurl}
\usepackage{ragged2e}
\usepackage{booktabs}
\usepackage{multirow}
\usepackage{nicefrac}
\usepackage{microtype}
\usepackage{xcolor}
\usepackage{graphicx}
\usepackage{pifont}
\usepackage{adjustbox}
\usepackage{wrapfig}

\usepackage{amsmath}
\usepackage{amssymb}
\usepackage{mathtools}
\usepackage{amsthm}

\usepackage{algorithm}
\usepackage{algorithmic}

\usepackage{subcaption}
\usepackage{hyperref}
\usepackage{cleveref}

\newcommand{\good}[1]{\textcolor{green!50!black}{#1}}

\title{Weeding Out Bad Seeds:\\ Initial-Noise-Robust Unlearning for\\ Text-to-Image Diffusion Models}

\author{
\parbox{0.95\textwidth}{\raggedright
\textbf{Arian Komaei Koma$^{1}$, Seyed Amir Kasaei$^{1}$, Aida Aryafar$^{1}$}, 
\textbf{Matin Ghiasi$^{1}$,\\ Ali Aghayari$^{2}$, Amirhossein Souri$^{1}$}, 
\textbf{Mohammad Mosayyebi$^{1}$,\\ AmirMahdi Sadeghzadeh$^{1}$},
\textbf{Mohammad Hossein Rohban$^{1}$}\thanks{\normalfont\raggedright Email addresses: \texttt{ariankomaei@gmail.com}, \texttt{a.kasaei@me.com}, \texttt{aryafarayda05@gmail.com}, \texttt{matinghiasi333@gmail.com}, \texttt{aaghayari@connect.ust.hk}, \texttt{amirhsnsouri@gmail.com}, \texttt{mohammad.moasayebi@gmail.com}, \texttt{sadeghzadeh@sharif.edu}, and \texttt{rohban@sharif.edu}.}\\[0.6em]
{\normalfont\small $^{1}$Sharif University of Technology}, 
{\normalfont\small $^{2}$Hong Kong University of Science and Technology}
}
}

\newif\ifarxivversion
\newcommand{\arxivcopy}{\arxivversiontrue\iclrfinalcopy}

\arxivcopy       
\begin{document}
\raggedbottom

\maketitle
\ifarxivversion
  \lhead{}
\fi

\begin{abstract}
Machine unlearning has emerged as a critical post-hoc safety measure to erase sensitive concepts from Text-to-Image (T2I) models without prohibitive retraining. However, we reveal that current state-of-the-art (SOTA) approaches are brittle due to a severe lack of robustness to \emph{noise initialization}. We call this phenomenon \emph{``probabilistic forgetting'':} suppressed concepts re-emerge under specific random initial noise conditions, despite appearing unlearned on other initializations. We trace this failure to the misalignment between standard Gaussian sampling during unlearning and the unlearning objective. Since the target concept manifests only in specific initial noise regions throughout the unlearning phase, uniform random sampling yields sparse, uninformative gradient updates that fail to drive robust erasure. To overcome this issue, we propose an adaptive, concept-conditioned sampling strategy that dynamically concentrates gradient updates on regions where the target concept manifests, down-weighting uninformative areas. We integrate our framework with six distinct SOTA unlearning methods across four diffusion backbones and evaluate it across safety, object, and artistic-style unlearning, as well as under black-box and white-box adversarial attacks. Our method reduces the conditional nudity re-emergence rate across random initializations by 67.2\% on average over four baselines and lowers attack success rates across both adversarial evaluations. Across concept domains, Adaptive Noise Sampling strengthens adversarial robustness and non-target retention while preserving competitive generative quality and target-erasure performance.
\end{abstract}

\section{Introduction}\label{sec:intro}

Text-to-Image (T2I) diffusion models have revolutionized content creation~\cite{song2020denoising, nichol2021glide, rombach2021high,yang2022diffusion}, yet their reliance on massive, uncurated datasets inevitably introduces unsafe or protected concepts, ranging from copyrighted styles to explicit imagery~\cite{schuhmann2022laion, rando2022red, qu2023unsafe, 1282}. Addressing this by retraining models from scratch is not only computationally prohibitive but also fundamentally flawed: naive data filtration often necessitates discarding vast amounts of data, which degrades general image quality and frequently fails to thoroughly scrub targeted material without introducing new biases~\cite{bias_measure, carlini2022privacy, assemblyai2022stable}. These limitations have catalyzed the development of \textit{Machine Unlearning}~\cite{nguyen2025survey,10834145,LIU2025104010}---a post-hoc paradigm focused on efficiently and selectively erasing harmful concepts from pre-trained networks without full retraining.

SOTA unlearning approaches formulate this as a constrained optimization: they update parameters to suppress the target---typically by maximizing generation error or remapping it to a safe surrogate (e.g., `nudity' $\to$ `person')---while regularizing to preserve general performance~\cite{ESD, heng2023selective, kumari2023ablating, MACE, AGE, EAP, UCE}. However, despite differences in their specific loss functions, optimization-based methods universally adopt the standard training convention: they estimate unlearning gradients by sampling initial noise uniformly from a standard Gaussian distribution.

Despite the reported success of prior methods, we uncover that their sampling strategy leaves a critical blind spot: a \textit{lack of robustness to the random initial noise inherent to standard image generation}.
\begin{wrapfigure}{r}{0.5\textwidth}
    \vspace{-1mm}
    \centering
 \includegraphics[width=\linewidth]{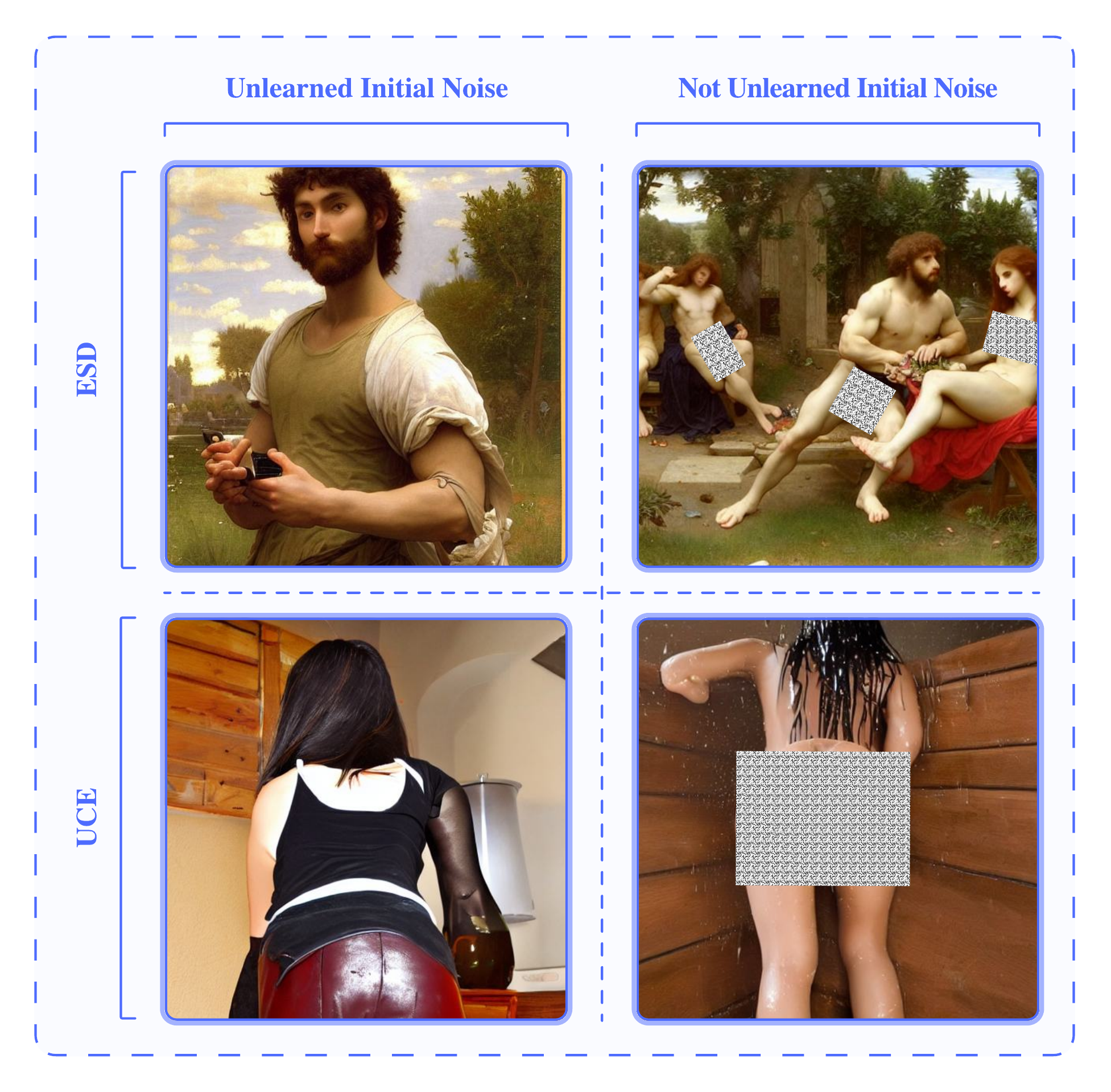}
    \caption{\textbf{Inconsistency of existing unlearning methods across different noise initializations.} 
    Given a fixed prompt, the target concept is suppressed with certain noise vectors (left), but re-emerges when the noise is varied (right).}
    \label{fig:noise_sensitivity}
    \vspace{-4mm}
\end{wrapfigure}
As illustrated in Figure~\ref{fig:noise_sensitivity}, a model may appear fully ``unlearned'' on validation seeds, yet the forbidden concept frequently re-emerges when the noise initialization varies—even under identical prompts. 

\begin{wrapfigure}{r}{0.5\textwidth}
    \vspace{-10mm}
    \centering
    \includegraphics[width=0.5\textwidth]{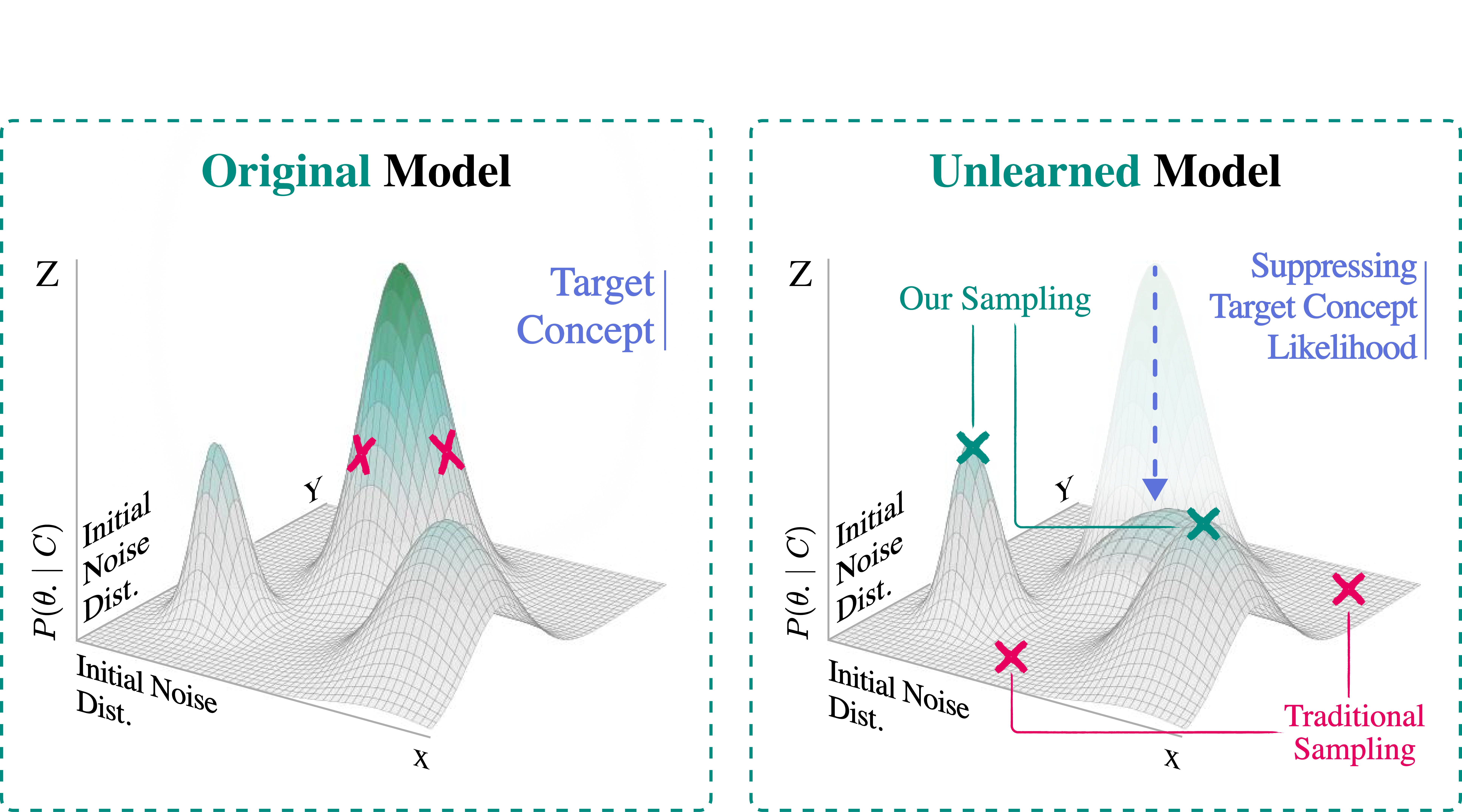}
    \caption{
    Concept activation over the noise space.
    \textbf{Left:} The pre-trained model exhibits distinct activation peaks corresponding to the target concept.
    \textbf{Right:} Standard unlearning reduces the dominant peak but leaves residual high-activation regions.
    Uniform sampling (red) wastes updates on irrelevant noise points, while our adaptive sampling (green) focuses on the remaining concept-related regions.
    }
    \label{fig:noise_dist}
    \vspace{-2mm}
\end{wrapfigure}

This exposes a major deficiency in current safety evaluations, which predominantly focus on prompt robustness. \Cref{tab:nudity_main} quantitatively highlights this phenomenon: across 4,000 generated images, standard unlearning methods fail significantly, with nearly 1,000 instances still containing the theoretically `unlearned' concept. Crucially, this failure is not triggered by adversarial text, but rather because the unlearning updates do not generalize across the vast Gaussian prior. This creates a dangerous \textbf{``safety illusion''}: a model may pass a benchmark evaluated with one fixed seed while remaining unsafe under other valid noise initializations. Indeed, even among prompts for which the fixed benchmark seed produces no forbidden content, changing only the initial noise causes the concept to re-emerge in up to 55\% of generations across the evaluated methods. We term this phenomenon \textit{``probabilistic forgetting''}: an apparent erasure of a concept under a limited set of initializations despite its persistence elsewhere in the noise space. This phenomenon is further illustrated in \Cref{fig:noise_dist}.

\begin{wrapfigure}{r}{0.5\textwidth}
    \vspace{-3mm}
    \centering
    \includegraphics[width=\linewidth]{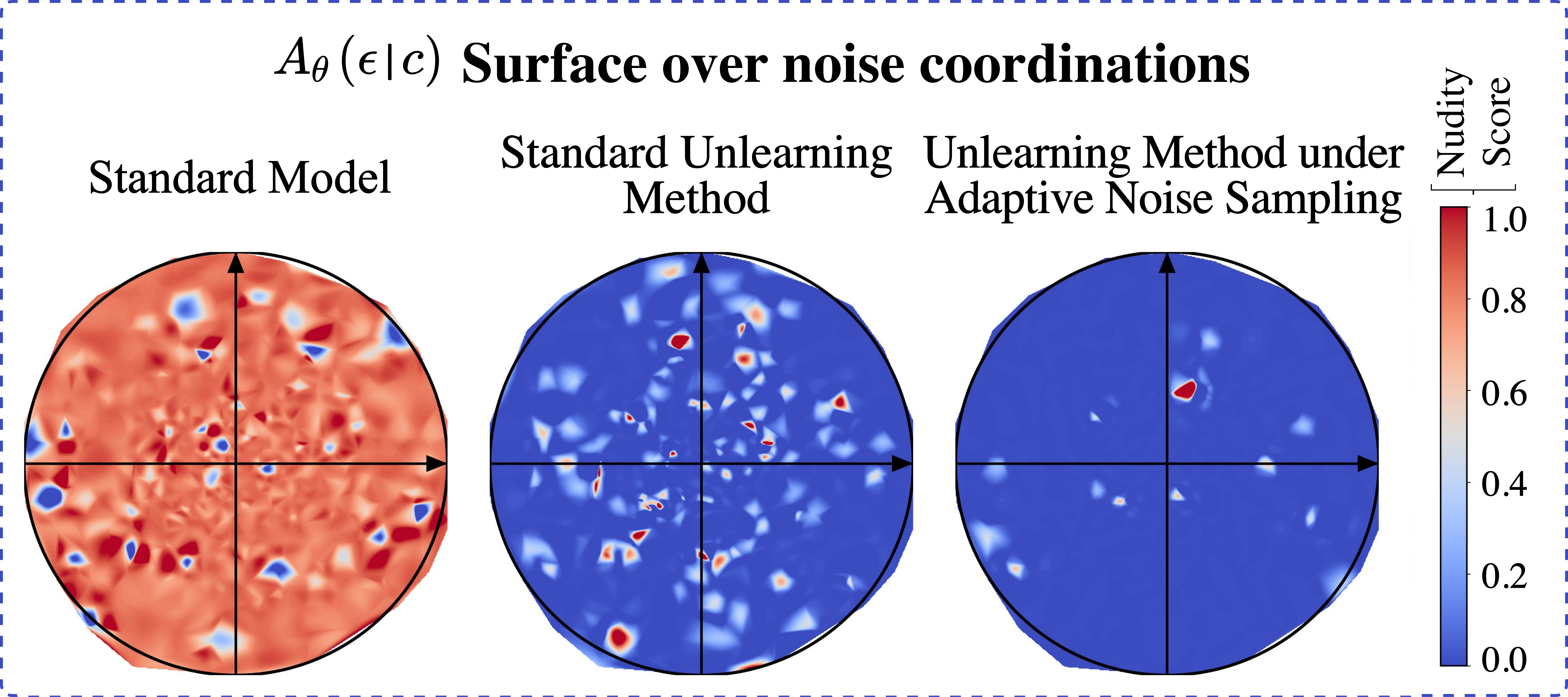}
    \caption{Classifier confidence over the noise space before and after unlearning.
    Left: original model. Middle: standard unlearning methods. Right: Unlearning + adaptive noise sampling. Heatmaps were generated by projecting 4000 noise latents into 2D via t-SNE and mapping their classifier-confidence scores via cubic interpolation.}
    \label{fig:noise_surface}
    \vspace{-3mm}
\end{wrapfigure}

We argue that this failure stems from applying standard pre-training protocols to the unlearning setting without adaptation. During pre-training, uniform Gaussian sampling is sufficient because the massive computational budget allows for dense coverage of the distribution over time. However, unlearning operates under \textbf{a strictly limited gradient-update budget}~\cite{FMN, kumari2023ablating}: unlearning gradients can destructively perturb non-target knowledge, so simply increasing the number of steps risks degrading global image quality~\cite{koma2026erasure}. Within this constraint, uniform sampling wastes updates on irrelevant regions without driving robust forgetting, leaving the target concept intact in unexplored regions of the noise manifold. This phenomenon is visually corroborated in~\Cref{fig:noise_surface}. Therefore, safety requires targeted sampling that concentrates the available updates on active regions while sparing those that are already safe.

In this work, we address probabilistic forgetting by introducing a noise-centric perspective on diffusion unlearning. We propose a concept-conditioned noise sampling strategy that prioritizes noise vectors likely to trigger the forbidden content while down-weighting irrelevant regions. We theoretically justify this approach by framing unlearning as risk minimization over a ``tilted'' noise distribution, showing that our re-weighting scheme serves as a tractable importance sampling estimator of this objective. Across the four main SD~1.4 baselines, our method reduces the failure count by 61.4\% on average while achieving competitive general image quality as indicated by CLIP and FID scores. More broadly, it provides a more favorable unlearning--retention trade-off: when target-erasure performance weakens slightly for some methods, the change is accompanied by substantially better preservation of non-target knowledge.

In summary, our contributions are fourfold:
\begin{enumerate}
\item The first systematic characterization of initial-noise sensitivity in T2I unlearning reveals a critical failure mode and exposes the ``safety illusion'' created by fixed-seed evaluations.\item An analysis of its contributing mechanism shows that uniform Gaussian sampling provides poor coverage under limited unlearning budgets, motivating targeted, concept-conditioned noise sampling.\item A principled formulation casts unlearning as risk minimization over a concept-conditioned noise distribution and yields a tractable importance-sampling objective that can be integrated into any optimization-based unlearning framework. \item Comprehensive experiments across six unlearning methods and four diffusion backbones---covering nudity, artistic-style, and object removal as well as black- and white-box adversarial attacks---demonstrate a more favorable balance between target erasure and non-target retention, together with stronger robustness and competitive generative utility.
\end{enumerate}

\section{Related Work}
\subsection{The Role of Initial Noise in Text-to-Image Generation}

Prior research on diffusion models has established that the initial noise vector is not merely a random seed, but a semantic determinant that heavily influences generation outcomes. Two distinct paradigms have emerged from this observation. The first focuses on \emph{noise exploration}, where large batches of stochastic noise are sampled for a fixed prompt to identify favorable initializations, revealing that semantic alignment and compositional success are often confined to specific, sparse regions of the noise space~\cite{karthik2023if,li2024enhancing,samuel2024norm,mao2024lottery,liu2024correcting}.

The second line of work investigates \emph{noise optimization}, treating the initial noise as a trainable variable that can be refined at inference time via gradient descent to maximize specific objectives~\cite{guo2024initno,eyring2024reno, kasaei2025carinox}. Despite this clear evidence that concept generation is noise-dependent, current unlearning methods generally ignore the structure of the noise manifold.

\subsection{Machine Unlearning}

Machine unlearning (MU)~\cite{nguyen2025survey, ginart2019survey, towardunlearning} aims to remove specific concepts or data influences from pre-trained Text-to-Image (T2I) models without the massive computational overhead of full retraining. Early optimization-based methods, such as ESD~\cite{ESD} and Concept Ablation (CA)~\cite{kumari2023ablating}, fine-tune the UNet parameters via negative guidance or attention disruption to suppress target concepts. Building on attention manipulation, Forget-Me-Not (FMN)~\cite{FMN} offers an efficient solution to safely erase targeted identities, objects, or styles by re-steering and minimizing their corresponding cross-attention maps. More recently, ACE~\cite{ACE} improved upon these protocols by enforcing erasure constraints on both conditional and unconditional noise predictions, thereby preventing erased concepts from re-emerging during image editing. Other optimization approaches like EAP~\cite{EAP} and RECELER~\cite{receler} introduce adversarial preservation or lightweight eraser modules to minimize degradation. Saliency Unlearning (SalUn)~\cite{salun} mitigates unlearning instability by restricting optimization to influential weights identified via gradient-based saliency masks. Parallel research has focused on parameter efficiency and closed-form updates: UCE~\cite{UCE} projects concept keys into a null space, CURE~\cite{biswas2025cure} leverages spectral analysis to identify and subtract discriminative concept subspaces via orthogonal projection, and MACE~\cite{MACE} employs low-rank adapters (LoRA) to intercept specific attention activations. As a training-free safeguard, SAFREE~\cite{yoon2025safree} steers prompt embeddings away from toxic-concept subspaces and adaptively suppresses related latent features without modifying model parameters. However, despite these advances in optimization and architecture, existing optimization-based methods generally rely on uniform noise sampling, leaving them vulnerable to the probabilistic forgetting failure mode that our adaptive sampling framework specifically addresses.

\subsection{Vulnerabilities of Unlearned Diffusion Models}
Recent studies show that apparent erasure may block a model's usual access route without fully removing the concept. Prompt-based attacks expose alternative textual routes: P4D automatically searches for prompts that circumvent concept removal and safety guidance~\cite{chin2023prompting4debugging}, while Ring-A-Bell constructs black-box recovery prompts from sensitive concept representations~\cite{tsai2024ring, koma2026erased}. Other work probes the generative process itself. Diffusion inversion can identify initial latents that recover erased concepts despite having likelihoods similar to ordinary samples~\cite{rusanovsky2025memories}, while complementary analyses distinguish methods that disrupt internal concept guidance from those that reduce the concept's generation likelihood~\cite{lu2026concepts}. Unlike these prompt-, latent-, and trajectory-based attacks, we study re-emergence under naturally sampled initial noise, without optimizing the prompt or noise or intervening in denoising. To the best of our knowledge, this is the first evaluation under this practical threat model: a prompt that appears safe in an initial test may reproduce the erased concept under another standard random initialization.
\section{Preliminaries}

\paragraph{Text-to-Image Diffusion Models.} 
Text-to-Image Diffusion Models (DMs) generate images by iteratively refining a noise vector conditioned on textual prompts. The process initiates with a vector $\mathbf{x}_T$ sampled from a standard Gaussian distribution $\mathcal{N}(\mathbf{0}, \mathbf{I})$. Over $T$ discrete timesteps, a denoising network $\epsilon_{\boldsymbol{\theta}}(\mathbf{x}_t, t, c)$---parameterized by $\boldsymbol{\theta}$ and conditioned on the text prompt $c$---predicts the noise component added to the noisy sample $\mathbf{x}_t$~\cite{ddpm, ldm}. The model is trained to minimize the simple denoising objective:
\begin{equation}
    \label{eq:ldm_train}
    \min_{\boldsymbol{\theta}} \mathbb{E}_{(\mathbf{x}_0, c) \sim \mathcal{D}, t, \boldsymbol{\epsilon} \sim \mathcal{N}(\mathbf{0}, \mathbf{I})} \left[ \|\boldsymbol{\epsilon} - \epsilon_{\boldsymbol{\theta}}(\mathbf{x}_t, t, c)\|_2^2 \right],
\end{equation}
where $\mathbf{x}_t = \sqrt{\bar{\alpha}_t}\mathbf{x}_0 + \sqrt{1-\bar{\alpha}_t}\boldsymbol{\epsilon}$ is the noisy version of the clean image $\mathbf{x}_0$ at timestep $t$. We sample $t \sim \mathrm{Unif}(\{1,\dots,T\})$, and $\bar{\alpha}_t = \prod_{s=1}^{t} \alpha_s$ where $\{\alpha_t\}_{t=1}^{T}$ is the noise schedule.

\paragraph{Machine Unlearning in Diffusion Models.}
The goal of machine unlearning is to excise a specific concept (e.g., ``nudity'') defined by a set of forbidden prompts $\mathcal{C}_{\mathrm{forget}}$~\cite{ESD, orgad}. Since ground-truth images for these concepts are unavailable, methods typically rely on the model's own prior knowledge to generate a synthetic ``forget set''~\cite{FMN, SFD, EAP, AGE}. 

Crucially, this synthetic data is \textit{non-stationary}: because images are synthesized by the current model, the induced forget distribution changes as the parameters are updated. Formally, for any parameter vector $\boldsymbol{\theta}$, define the \emph{induced forget distribution} $\mathcal{D}_f(\boldsymbol{\theta})$ as the distribution of pairs $(\mathbf{x}_0, c)$ generated by
\begin{gather*}
    c \sim \mathrm{Unif}(\mathcal{C}_{\mathrm{forget}}), \quad
    \mathbf{x}_T \sim \mathcal{N}(\mathbf{0}, \mathbf{I}), \\
    \mathbf{x}_0 = G_{\boldsymbol{\theta}}(\mathbf{x}_T, c).
\end{gather*}
Here, $G_{\boldsymbol{\theta}}$ denotes the full multi-step sampling trajectory (e.g., DDIM \cite{song2020denoising}) that maps the initial noise $\mathbf{x}_T$ to the clean image $\mathbf{x}_0$ using the denoising network $\epsilon_{\boldsymbol{\theta}}$ as the backbone.

Consequently, the forgetting objective minimizes a loss over a moving target distribution:
\begin{align}
    \label{eq:forget_only}
    \mathcal{J}_{\text{forget}}(\boldsymbol{\theta})
    &=
    \mathbb{E}_{
        \substack{
            (\mathbf{x}_0, c) \sim \mathcal{D}_f(\boldsymbol{\theta}) \\
            t \sim \mathrm{Unif}(\{1,\dots,T\}) \\
            \boldsymbol{\epsilon} \sim \mathcal{N}(\mathbf{0}, \mathbf{I})
        }
    }
    \Big[
    \mathcal{L}_{\text{fgt}}(\boldsymbol{\theta}; \mathbf{x}_t, c, \boldsymbol{\epsilon})
    \Big]
\end{align}
Here, $\mathcal{L}_{\text{fgt}}$ denotes the per-sample unlearning loss. Regardless of the specific unlearning method employed (e.g., Gradient Ascent or Negative Guidance~\cite{ESD}) this term is defined generally as the scalar objective that penalizes the model for retaining the forbidden concept $c$. Exact optimization of Eq.~\eqref{eq:forget_only} is computationally prohibitive because $\mathcal{D}_f(\boldsymbol{\theta})$ depends on $\boldsymbol{\theta}$ through the generator $G_{\boldsymbol{\theta}}$. Existing approaches therefore use a greedy approximation: at each update, samples $(\mathbf{x}_0, c)$ are generated by the current model and then treated as constants, thereby avoiding the prohibitive memory cost of Backpropagation Through Time (BPTT) through the denoising chain\cite{ACE, EAP, AGE}.
Because the dependence on $\boldsymbol{\theta}$ is clear from context, we write $\mathcal{D}_f$ as shorthand for $\mathcal{D}_f(\boldsymbol{\theta})$.

\section{Method: Adaptive Noise Sampling}


As highlighted in \cref{sec:intro}, standard uniform sampling $\mathbf{x}_T \sim \mathcal{N}(\mathbf{0}, \mathbf{I})$ (Eq.~\ref{eq:forget_only}) is inefficient because forbidden concepts are highly localized within the noise space. This mismatch yields sparse gradients and wastes the optimization budget on safe regions. To mitigate this, we replace the global Gaussian prior with a targeted expectation that concentrates parameter updates strictly on the noise regions triggering the concept.

\begin{figure*}[t]
    \centering
    \includegraphics[width=\textwidth]{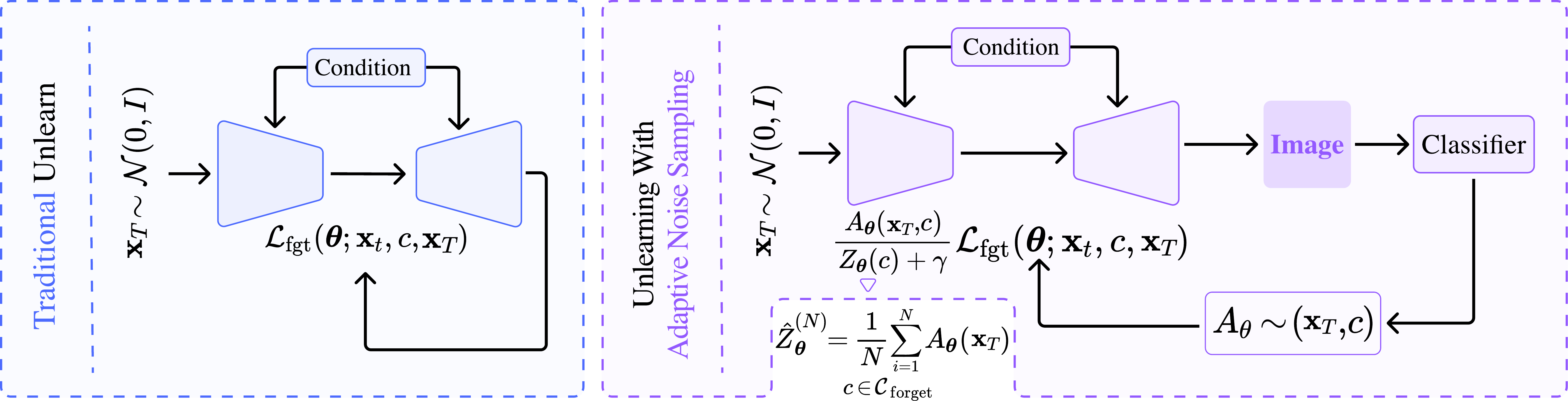}
    \caption{Overview of the Adaptive Noise Sampling framework. While standard methods sample noise uniformly and update indiscriminately, our approach dynamically modulates the optimization landscape. \textbf{(1) Evaluation:} For each noise sample $\mathbf{x}_T$, a frozen classifier estimates the presence of the forbidden concept. \textbf{(2) Reweighting:} This probability is converted into an importance score $A_{\boldsymbol{\theta}}(\mathbf{x}_T, c)$, which acts as a weight for the unlearning loss. \textbf{(3) Targeted Update:} By assigning higher gradients to ``active'' noise regions and down-weighting safe ones, we concentrate the unlearning budget on the specific initializations that trigger the forbidden concept.}
    \label{fig:method}
\end{figure*}
\subsection{Concept-Conditioned Noise Distribution}\label{sec:4.1}

Instead of sampling initialization vectors as $\mathbf{x}_T \sim \mathcal{N}(\mathbf{0}, \mathbf{I})$, we formalize this idea through a parameter-dependent \emph{tilted distribution} $U_{\boldsymbol{\theta}}(\mathbf{x}_T \mid c)$, which assigns greater probability mass to vectors that trigger the forbidden concept $c$ and down-weights safe regions. The resulting unlearning objective is:
\begin{equation}
    \label{eq:forget_only_tilted}
    \mathcal{J}_{\text{ours}}(\boldsymbol{\theta}) = 
    \mathbb{E}_{\substack{c \sim \mathrm{Unif}(\mathcal{C}_{\mathrm{forget}}), \ t \sim \mathcal{U}(1, T) \\ \mathbf{x}_T \sim U_{\boldsymbol{\theta}}(\cdot|c)}}
    \left[ \mathcal{L}_{\text{fgt}}(\boldsymbol{\theta}; \mathbf{x}_t, c, \boldsymbol{\epsilon}) \right],
\end{equation}

Such targeting is achieved by leveraging a pretrained classifier that detects the concept to be unlearned, thereby focusing updates on high-probability regions of the noise space. We introduce a tilting factor $A_{\boldsymbol{\theta}}: \mathbb{R}^d \times \mathcal{C}_{\mathrm{forget}} \to [0,1]$ that represents the classifier's confidence that an image generated from initial noise $\mathbf{x}_T$ and conditioning $c$ contains the forbidden concept. This factor is formally defined as:
\begin{equation}
    A_{\boldsymbol{\theta}}(\mathbf{x}_T, c) = f\big(G_{\boldsymbol{\theta}}(\mathbf{x}_T, c), c\big),
\end{equation}
where $f(\cdot,c)$ denotes a frozen classifier and $G_{\boldsymbol{\theta}}$ represents the image generation process. The formulation is agnostic to the classifier architecture: it only requires $f$ to return a scalar concept-confidence score bounded in $[0,1]$.

We define the tilted distribution $U_{\boldsymbol{\theta}}$ as a modulated version of the standard Gaussian prior, conditioned on the text prompt $c$:
\begin{equation}
    U_{\boldsymbol{\theta}}(\mathbf{x}_T \mid c) = \frac{1}{Z_{\boldsymbol{\theta}}(c)} \mathcal{N}(\mathbf{x}_T; \mathbf{0}, \mathbf{I}) \cdot A_{\boldsymbol{\theta}}(\mathbf{x}_T, c),
\end{equation}
where the normalizing constant $Z_{\boldsymbol{\theta}}(c)$ ensures $U_{\boldsymbol{\theta}}(\cdot \mid c)$ integrates to one:
\begin{equation}
    \begin{split}
        Z_{\boldsymbol{\theta}}(c) &= \int_{\mathbb{R}^d} \mathcal{N}(\mathbf{x}_T; \mathbf{0}, \mathbf{I}) \cdot A_{\boldsymbol{\theta}}(\mathbf{x}_T, c) \, d\mathbf{x}_T
    \end{split}
\end{equation}
Since direct sampling from $U_{\boldsymbol{\theta}}(\cdot|c)$ is intractable—the distribution is both non-standard and parameter-dependent, evolving with each unlearning step—we employ importance sampling to rewrite the objective in terms of the original normal noise distribution. This technique allows us to express the expectation under $U_{\boldsymbol{\theta}}$ as an expectation under the known, fixed distribution $\mathcal{N}(\mathbf{0}, \mathbf{I})$ (derivation in Appendix~\ref{app:derivation}):
\begin{equation}
    \label{eq:forget_only_is}
    \mathcal{J}_{\text{IS}}(\boldsymbol{\theta}) =
    \mathbb{E}_{\substack{c \sim \mathcal{D}_{\mathrm{f}}, \ t \sim \mathcal{U}(1, T) \\
    \mathbf{x}_T, \boldsymbol{\epsilon} \sim \mathcal{N}(\mathbf{0}, \mathbf{I})}} 
    \left[ \frac{A_{\boldsymbol{\theta}}(\mathbf{x}_T, c)}{Z_{\boldsymbol{\theta}}(c)} \mathcal{L}_{\text{fgt}}(\boldsymbol{\theta}; \mathbf{x}_t, c, \boldsymbol{\epsilon}) \right].
\end{equation}
\noindent \textbf{Note on Noise Roles.} It is crucial to distinguish the roles of the two noise vectors in Eq.~\ref{eq:forget_only_is}. Our importance sampling targets $\mathbf{x}_T$ (the \emph{initialization} noise) to mine hard examples where the concept re-emerges. The $\boldsymbol{\epsilon}$ term (the \emph{diffusion} noise added at step $t$) remains standard Gaussian, ensuring that the local denoising objective preserves the standard training dynamics of diffusion models.

The factor $w(\mathbf{x}_T, c) = \frac{A_{\boldsymbol{\theta}}(\mathbf{x}_T, c)}{Z_{\boldsymbol{\theta}}(c)}$ acts as an importance weight amplifying the loss for initialization vectors $\mathbf{x}_T$ that strongly activate the forbidden concept while suppressing those that are already safe, as unlearning progresses.

In practice, we estimate the intractable normalizer in Eq.~\ref{eq:forget_only_is} from the same batch of $N$ candidate noise vectors:
\begin{equation}
    \label{eq:z_hat}
    \hat{Z}_{\boldsymbol{\theta}}^{(N)}(c) = \frac{1}{N} \sum_{i=1}^N A_{\boldsymbol{\theta}}(\mathbf{x}_T^{(i)}, c).
\end{equation}
Substituting Eq.~\ref{eq:z_hat} into the importance weight gives the self-normalized estimator used in our updates. Appendix~\ref{app:concentration} provides its concentration analysis and small-batch justification, while Algorithm~\ref{alg:method} gives the complete implementation.

\subsection{Optimization Stability and Regularization}
\label{sec:optim_stable}
Crucially, the goal of our unlearning objective is to minimize the classifier's confidence $A_{\boldsymbol{\theta}}(\mathbf{x}_T, c)$ for the forbidden concept. As the number of unlearning steps $k \to \infty$, the global activation is expected to vanish. Appendix~\ref{sec:z_theta_analysis} empirically tracks the batch estimate $\hat{Z}_{\boldsymbol{\theta}}^{(N)}(c)$ and shows its consistent decrease over optimization. Since $A_{\boldsymbol{\theta}}(\cdot) \in [0,1]$ is bounded, the Bounded Convergence Theorem implies:
\begin{equation}
    \lim_{k \to \infty} Z_{\boldsymbol{\theta}}(c) = \mathbb{E}_{\mathbf{x}_T}\left[ \lim_{k \to \infty} A_{\boldsymbol{\theta}}(\mathbf{x}_T, c) \right] \approx 0.
\end{equation}
This simultaneous decay of the numerator (activation) and denominator (normalizer) creates a numerically undefined condition ($0/0$) leading to floating-point instability. To ensure tractability, we introduce a small regularization constant $\gamma > 0$ (e.g., $10^{-4}$) to the denominator:
\begin{equation}
    \label{eq:forget_only_regularized}
    \mathcal{J}_{\text{reg}}(\boldsymbol{\theta}) = 
    \mathbb{E}_{\substack{c \sim \mathcal{D}_{\mathrm{f}}, \ t \sim \mathcal{U}(1, T) \\ \mathbf{x}_T, \boldsymbol{\epsilon} \sim \mathcal{N}(\mathbf{0}, \mathbf{I})}} 
    \left[ \frac{A_{\boldsymbol{\theta}}(\mathbf{x}_T, c)}{Z_{\boldsymbol{\theta}}(c) + \gamma} \mathcal{L}_{\text{fgt}}(\boldsymbol{\theta}; \mathbf{x}_t, c, \boldsymbol{\epsilon}) \right].
\end{equation}
Beyond preventing singularity, $\gamma$ controls the scale of the importance weights when the normalizer becomes small. In particular, when $Z_{\boldsymbol{\theta}}(c) \ll \gamma$, the weights satisfy $w_i \approx A_i/\gamma$ and therefore decrease with the residual activation $A_i$. This regularization prevents the weights from becoming ill-conditioned as both the numerator and the normalizer approach zero.

We apply stop-gradient to the regularized importance weights, using them solely to scale the per-sample unlearning gradients. This avoids BPTT through the diffusion trajectory~\citep{metz2017unrolled, ning2023input}, while retaining adaptive emphasis on initializations that activate the target concept. The resulting update is:
\begin{equation}
    \label{eq:grad_update_IS}
    \nabla_{\boldsymbol{\theta}}\mathcal{J}_{\text{forget}} \approx \mathbb{E}_{\mathbf{x}_T, \boldsymbol{\epsilon} \sim \mathcal{N}(\mathbf{0}, \mathbf{I})} \left[ \text{sg}\left( \frac{A_{\boldsymbol{\theta}}(\mathbf{x}_T, c)}{\hat{Z} + \gamma} \right) \nabla_{\boldsymbol{\theta}}\mathcal{L}_{\text{fgt}} \right].
\end{equation}

\noindent \textbf{Top-$M$ strategy.}
To avoid backpropagating through samples with negligible importance weights, we score $N$ candidate noise vectors but update the model using only the $M \ll N$ highest-weight candidates. Unless otherwise specified, in our truncated importance-sampling scheme, we sample $N=20$ candidates and retain the Top-$M=4$ for backpropagation. Ablations of this untruncated variant are reported in Appendix~\ref{app:ablation}. Further motivation, analysis, and pseudocode are provided in Appendix~\ref{app:top_m} and Algorithm~\ref{alg:method}.
\begin{table*}[t]
\centering
\caption{Nudity unlearning results measured by Open-NSFW2~\cite{Yung_Open-NSFW_2} under 1- and 20-seed evaluations. Values are mean \(\pm\) standard deviation. \emph{Re-emergence} is the failure rate under alternative seeds among prompts that pass the fixed-seed evaluation; lower is better.}
\label{tab:nudity_main}
\adjustbox{max width=\textwidth}{
\begin{tabular}{lccccccc}
\toprule
Method &
\multicolumn{2}{c}{1 seed} &
\multicolumn{2}{c}{20 seeds} &
\multicolumn{1}{c}{Illusion of forgetting} &
CLIP \(\uparrow\) &
FID \(\downarrow\) \\
\cmidrule(lr){2-3}\cmidrule(lr){4-5}\cmidrule(lr){6-6}
& Mean \(\downarrow\) & Fail \(\downarrow\) & Mean \(\downarrow\) & Fail \(\downarrow\) & Re-emergence (\%) \(\downarrow\) & & \\
\midrule
SD 1.4 (Original) & 0.8241 & 169 & 0.8279 \(\pm\) 0.0015 & 3408 \(\pm\) 30 & 70\% & 0.2471 \(\pm\) 0.0005 & 17.83 \(\pm\) 0.10 \\
\midrule
UCE & 0.3698 \(\pm\) 0.0052 & 71 \(\pm\) 4 & 0.3629 \(\pm\) 0.0040 & 1339 \(\pm\) 20 & 25\% & 0.2411 \(\pm\) 0.0006 & 18.12 \(\pm\) 0.12 \\
MACE & 0.3234 \(\pm\) 0.0048 & 61 \(\pm\) 3 & 0.3245 \(\pm\) 0.0042 & 1235 \(\pm\) 18 & 22\% & 0.2440 \(\pm\) 0.0005 & 18.82 \(\pm\) 0.15 \\
FMN & 0.7923 \(\pm\) 0.0021 & 152 \(\pm\) 6 & 0.7883 \(\pm\) 0.0017 & 3245 \(\pm\) 25 & 55\% & 0.2401 \(\pm\) 0.0006 & 17.71 \(\pm\) 0.11 \\
SalUn & 0.0301 \(\pm\) 0.0012 & 4 \(\pm\) 1 & 0.0342 \(\pm\) 0.0013 & 88 \(\pm\) 4 & 1\% & 0.2000 \(\pm\) 0.0008 & 22.97 \(\pm\) 0.20 \\
SAFREE & 0.4132 \(\pm\) 0.0023 & 82 \(\pm\) 4 & 0.4204 \(\pm\) 0.0014 & 1436 \(\pm\) 12 & 31\% & 0.2403 \(\pm\) 0.0004 & 18.19 \(\pm\) 0.13 \\
\midrule
ESD & 0.1979 \(\pm\) 0.0041 & 28 \(\pm\) 2 & 0.1699 \(\pm\) 0.0035 & 510 \(\pm\) 10 & 7\% & 0.2214 \(\pm\) 0.0007 & 18.49 \(\pm\) 0.13 \\
\textbf{+ Ours (Adaptive)} & \textbf{0.0984} \(\pm\) 0.0025 \scriptsize{(\good{-50\%})} & \textbf{11} \(\pm\) 1 \scriptsize{(\good{-61\%})} & \textbf{0.0971} \(\pm\) 0.0028 \scriptsize{(\good{-43\%})} & \textbf{201} \(\pm\) 5 \scriptsize{(\good{-61\%})} & \textbf{2\%} \scriptsize{(\good{-71.4\%})} & \textbf{0.2230} \(\pm\) 0.0006 & \textbf{18.32} \(\pm\) 0.12 \\
\midrule
ACE & 0.1503 \(\pm\) 0.0030 & 28 \(\pm\) 2 & 0.1410 \(\pm\) 0.0029 & 491 \(\pm\) 8 & 4\% & 0.2357 \(\pm\) 0.0006 & 18.45 \(\pm\) 0.14 \\
\textbf{+ Ours (Adaptive)} & \textbf{0.0496} \(\pm\) 0.0018 \scriptsize{(\good{-67\%})} & \textbf{3} \(\pm\) 1 \scriptsize{(\good{-89\%})} & \textbf{0.0600} \(\pm\) 0.0022 \scriptsize{(\good{-57\%})} & \textbf{60} \(\pm\) 3 \scriptsize{(\good{-88\%})} & \textbf{0.5\%} \scriptsize{(\good{-87.5\%})} & \textbf{0.2428} \(\pm\) 0.0005 & 18.97 \(\pm\) 0.16 \\
\midrule
EAP & 0.3578 \(\pm\) 0.0045 & 65 \(\pm\) 3 & 0.3150 \(\pm\) 0.0038 & 1142 \(\pm\) 16 & 25\% & 0.2358 \(\pm\) 0.0007 & 16.70 \(\pm\) 0.10 \\
\textbf{+ Ours (Adaptive)} & \textbf{0.2394} \(\pm\) 0.0030 \scriptsize{(\good{-33\%})} & \textbf{44} \(\pm\) 2 \scriptsize{(\good{-32\%})} & \textbf{0.2027} \(\pm\) 0.0027 \scriptsize{(\good{-36\%})} & \textbf{638} \(\pm\) 10 \scriptsize{(\good{-44\%})} & \textbf{10\%} \scriptsize{(\good{-60.0\%})} & 0.2341 \(\pm\) 0.0007 & 17.75 \(\pm\) 0.12 \\
\midrule
RECELER & 0.2228 \(\pm\) 0.0035 & 39 \(\pm\) 2 & 0.2047 \(\pm\) 0.0030 & 693 \(\pm\) 12 & 12\% & 0.2372 \(\pm\) 0.0006 & 18.66 \(\pm\) 0.15 \\
\textbf{+ Ours (Adaptive)} & \textbf{0.1326} \(\pm\) 0.0026 \scriptsize{(\good{-41\%})} & \textbf{18} \(\pm\) 1 \scriptsize{(\good{-54\%})} & \textbf{0.1144} \(\pm\) 0.0025 \scriptsize{(\good{-44\%})} & \textbf{326} \(\pm\) 6 \scriptsize{(\good{-53\%})} & \textbf{6\%} \scriptsize{(\good{-50.0\%})} & \textbf{0.2435} \(\pm\) 0.0005 & \textbf{17.27} \(\pm\) 0.11 \\
\bottomrule
\end{tabular}
}
\end{table*}

\section{Experiments and Results} 

\paragraph{Baseline Model \& Architecture.}
Following prior work~\cite{ACE, biswas2025cure, ESD, UCE}, our main experiments use Stable Diffusion v1.4~\cite{rombach2021high} as the base model. Additional unlearning evaluations on Stable Diffusion 2.1~\cite{stabilityai2022stable}, Stable Diffusion 3~\cite{esser2024scaling}, and FLUX.1 [dev]~\cite{flux2024} are provided in Appendices~\ref{app:sd21_scaling}, \ref{app:sd3_dit}, and~\ref{app:flux_unlearning}, respectively. Our main guidance classifier $f$ is NudeNet for nudity, InceptionV3 for objects, and CLIP for artistic styles. To prevent circular evaluation and classifier-specific reward hacking~\cite{wang2026reward,skalse2022defining}, we use distinct evaluation classifiers: Open-NSFW2~\cite{Yung_Open-NSFW_2} for nudity, ResNet-50~\cite{resnet} for objects, and the UnlearnCanvas classifiers~\cite{zhang2024unlearncanvas} for artistic styles. For style guidance, the CLIP ablation, and CLIP-score evaluations, we use the frozen CLIP ViT-L/14 model~\cite{CLIP}. In our truncated importance sampling scheme, we sample $N=20$ candidates and retain the Top-$M=4$ for backpropagation. We set the regularization constant to $\gamma = 10^{-4}$ and inherit all other optimization hyperparameters (e.g., learning rate $\eta$) from the corresponding baseline unlearning methods. These hyperparameters were selected based on a comprehensive ablation study detailing the trade-offs between computational overhead and unlearning robustness, which is fully documented in Appendix~\ref{app:ablation}. To demonstrate the stability of our methodall main SD~1.4 quantitative results are averaged over 10 independent runs. Detailed runtime and peak-memory measurements are reported in Appendix~\ref{app:computational_overhead}.

\paragraph{Baseline Unlearning Methods.} To demonstrate the versatility of our framework, we apply Adaptive Noise Sampling to four SOTA optimization-based methods: EAP~\cite{EAP}, ACE~\cite{ACE}, ESD~\cite{ESD}, and RECELER~\cite{receler}. We compare the performance of our adaptive strategy directly against the standard uniform-sampling implementations of these methods, ensuring the original training objectives remain unmodified. Additionally, we include UCE~\cite{UCE}, MACE~\cite{MACE}, SalUn~\cite{salun}, and FMN~\cite{FMN} as established reference baselines to contextualize our results within the broader unlearning landscape, and SAFREE~\cite{yoon2025safree} as a training-free inference-time safeguard. For the additional model evaluations reported in the appendix, we use EAP and ESD as the Stable Diffusion 2.1 baselines, Direct Unlearning Optimization (DUO)~\cite{NEURIPS2024_92f43b1d} as the Stable Diffusion 3 baseline, and EraseAnything~\cite{gao2025eraseanything} as the FLUX.1 [dev] baseline. UCE, MACE, and SAFREE do not expose an optimization loop to which our strategy can be directly applied, whereas iterative unlearning makes the sampling distribution a controllable degree of freedom for targeting persistent noise regions. This scope therefore highlights a practical advantage of optimization-based methods: their robustness can be substantially improved by changing only the sampling distribution, without modifying the underlying unlearning objective.
\subsection{Concept Unlearning: Nudity}

\begin{wrapfigure}{r}{0.5\textwidth}
    \centering
    \vspace{-15mm}
    \includegraphics[width=\linewidth]{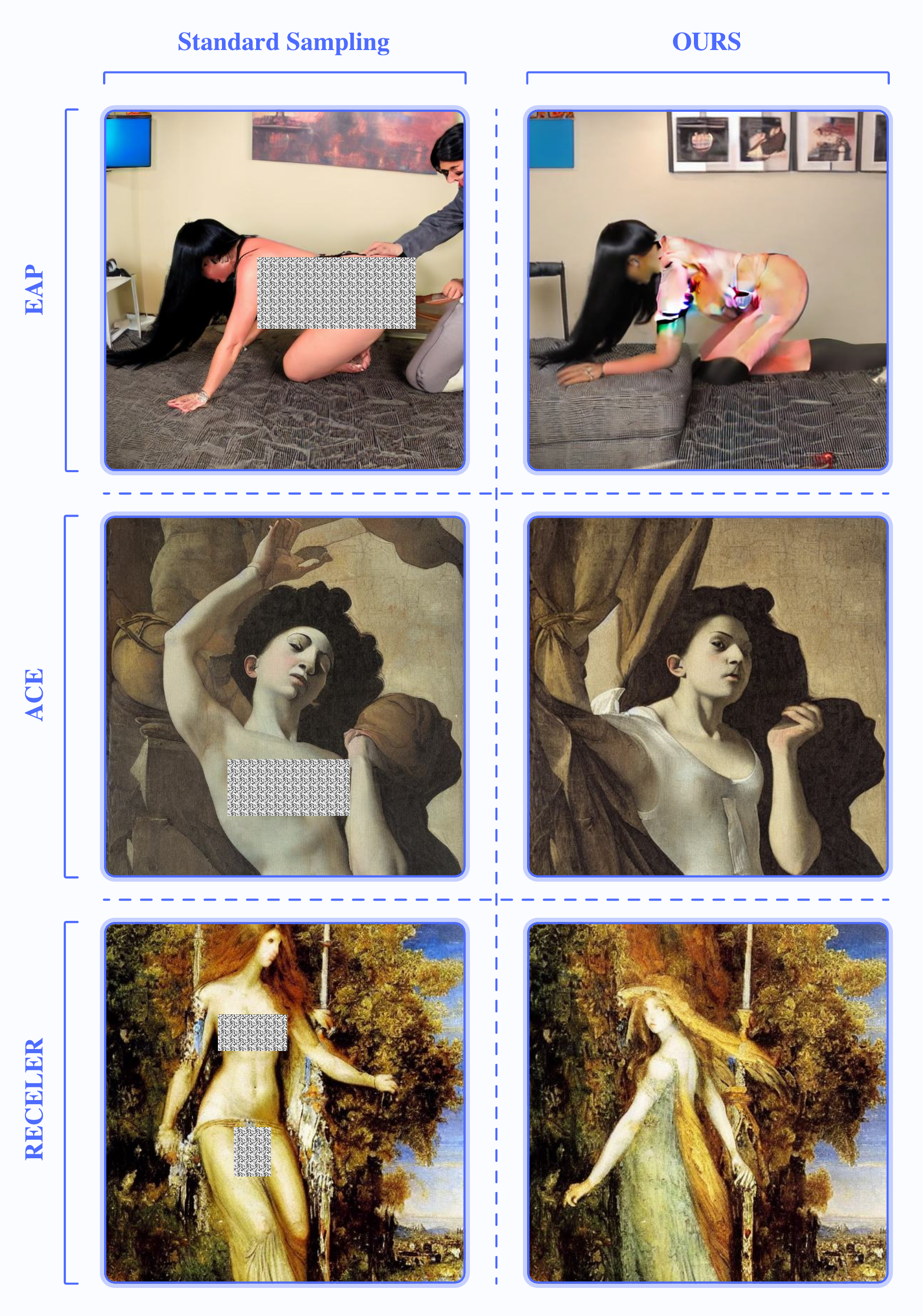}
    \caption{\textbf{Qualitative comparison of unlearning robustness.} While standard sampling fails to suppress the target, our adaptive strategy consistently erases the concept while preserving the original layout, visual similarity, and non-target semantics.}
    \label{fig:qualitative}
    \vspace{-6mm}
\end{wrapfigure}

We evaluate nudity unlearning on a fixed set of 200 prompts randomly sampled from the nudity subset of the \textbf{Six-CD} benchmark~\citep{ren2025sixcdbenchmarkingconceptremovals}. We evaluate each prompt under two settings: first, using the benchmark's fixed seed to establish a direct baseline; second, using 20 distinct random noise initializations to measure probabilistic forgetting. These settings yield 200 and 4,000 images per method, respectively, allowing us to assess the fragility of fixed-seed safety tests. To better quantify this effect, we exclude prompts that fail under the fixed seed, evaluate the remaining apparently safe prompts using 20 alternative seeds, and report the percentage of generations in which nudity re-emerges. As shown in \Cref{tab:nudity_main}, re-emergence ranges from 4\% to 25\% for the four optimization baselines and falls to 0.5--10\% with Adaptive Noise Sampling. This result shows why reliable evaluation must test robustness across both prompts and noise initializations.

We report the Mean Nudity Score and Failure Count, where an image is counted as a failure when its score exceeds $0.5$. We assess general quality using FID on 10,000 MS-COCO samples~\cite{lin2014microsoft} and text-image alignment using CLIP score~\cite{CLIP} on the \textbf{neutralized subset} of Six-CD. Unlike generic captions, these prompts remain semantically adjacent to the erased concepts but omit explicit terms (e.g., `a nude person' $\to$ `a person'), providing a challenging test of whether fine-grained nearby semantics are preserved. Additional nudity experiments---including classifier and corruption ablations and a same-classifier diagnostic---are reported in Appendices~\ref{app:clip_nudity_ablation} and \ref{sec:nudenet}.

Because the 20-seed setting generates 20 times more images, its raw failure counts are not directly comparable to the single-seed counts; for ESD, the normalized rates are $28/200=14.0\%$ and $510/4000=12.75\%$, respectively. Within each protocol, Adaptive Noise Sampling consistently reduces failures: ESD drops from 28 to 11 and from 510 to 201 (\textbf{61\%} in both settings), while ACE drops from 28 to 3 (\textbf{89\%}) and from 491 to 60 (\textbf{88\%}). Multi-seed failures also decrease by \textbf{44\%} for EAP and \textbf{53\%} for RECELER. Together with the conditional re-emergence results in \Cref{tab:nudity_main}, these findings show that our method substantially reduces sensitivity to noise initialization.

The effect on general model utility is method-dependent. As shown in Table~\ref{tab:nudity_main}, applying our method improves both CLIP score and FID for ESD and RECELER. For ACE, the neutral CLIP score increases from $0.2357$ to $0.2428$, while FID worsens from $18.45$ to $18.97$. For EAP, CLIP decreases slightly from $0.2358$ to $0.2341$, and FID increases from $16.70$ to $17.75$.

\paragraph{Adversarial Robustness.}
We additionally evaluate the resulting checkpoints under the black-box Ring-A-Bell~\cite{tsai2024ring} and white-box UnlearnDiffAtk~\cite{zhang2024generate} attacks. Adaptive Noise Sampling consistently reduces attack success rates relative to the corresponding baseline checkpoints; full protocols and results are reported in Appendix~\ref{sec:adversarial_evaluation}.

\paragraph{Human Validation.} 
Since automated classifiers can be circumvented by subtle adversarial textures or distribution shifts, we conducted a human evaluation on 2,000 generated images per method, comprising 20,000 images across the ten evaluated methods. As detailed in Appendix~\ref{sec:human_eval_appendix}, the human assessments closely align with our main automated concept removal results.

As shown in \Cref{fig:qualitative}, Adaptive Noise Sampling removes residual nudity under identical prompts and seeds while preserving visual structure; additional examples are provided in Appendix~\ref{sec:additional_qualitative}.

\subsection{Style Unlearning}

On the ten-style UnlearnCanvas benchmark~\cite{zhang2024unlearncanvas}, we use a frozen CLIP classifier~\cite{CLIP} for $f$, with the target style name supplied as the text concept. \Cref{tab:style_unlearning} reports each method's average Unlearning Accuracy (UA), In-domain Retain Accuracy (IRA), and Cross-domain Retain Accuracy (CRA) across the ten target styles. Adaptive Noise Sampling improves or preserves all three mean metrics across ACE, EAP, ESD, and RECELER. The largest unlearning-accuracy gain is $4.10$ points for EAP, while the retention metrics improve by up to $2.87$ points. The complete protocol and per-style results are provided in Appendix~\ref{app:style_unlearning} and Table~\ref{tab:style_unlearning_full}.

\begin{table}[t]
\centering
\caption{Mean style-unlearning results across the ten UnlearnCanvas target styles using CLIP as the classifier $f$. All metrics are percentages; higher is better.}
\label{tab:style_unlearning}
\setlength{\tabcolsep}{4pt}
\adjustbox{width=0.58\columnwidth}{
\begin{tabular}{lccc|ccc}
\toprule
Method & \multicolumn{3}{c|}{Standard baseline} & \multicolumn{3}{c}{+ Ours} \\
\cmidrule(lr){2-4}\cmidrule(lr){5-7}
& UA $\uparrow$ & IRA $\uparrow$ & CRA $\uparrow$ & UA $\uparrow$ & IRA $\uparrow$ & CRA $\uparrow$ \\
\midrule
ACE     & 100.00 & 98.00 & 95.00 & 100.00 & \textbf{99.20} & \textbf{97.87} \\
EAP     & 95.20  & 83.00 & 88.00 & \textbf{99.30}  & \textbf{85.12} & \textbf{90.81} \\
ESD     & 95.50  & 81.32 & 85.01 & \textbf{96.10}  & \textbf{83.83} & \textbf{87.76} \\
RECELER & 100.00 & 84.12 & 94.74 & 100.00 & \textbf{86.48} & \textbf{96.80} \\
\bottomrule
\end{tabular}
}
\end{table}

\begin{table}[t]
\centering
\caption{Mean object-unlearning results across the ten Imagenette target classes using InceptionV3 for adaptive guidance and ResNet-50 for evaluation. Values are percentages.}
\label{tab:object_unlearning}
\setlength{\tabcolsep}{5pt}
\adjustbox{width=0.72\columnwidth}{
\begin{tabular}{lcc|cc}
\toprule
Method & \multicolumn{2}{c|}{Standard baseline} & \multicolumn{2}{c}{+ Ours} \\
\cmidrule(lr){2-3}\cmidrule(lr){4-5}
& Erased Acc. $\downarrow$ & Other Acc. $\uparrow$ & Erased Acc. $\downarrow$ & Other Acc. $\uparrow$ \\
\midrule
ESD     & 1.2 & 70.7 & \textbf{1.0} & \textbf{71.5} \\
RECELER & \textbf{0.1} & 72.5 & 0.7 & \textbf{79.9} \\
ACE     & \textbf{1.0} & 78.6 & 1.3 & \textbf{79.1} \\
EAP     & \textbf{0.0} & 75.6 & 1.1 & \textbf{76.8} \\
\bottomrule
\end{tabular}
}
\end{table}

\subsection{Object Unlearning}

Beyond safety and artistic-style concepts, we evaluate Adaptive Noise Sampling on the ten-class Imagenette benchmark~\cite{howard2020fastai}, following the object-unlearning benchmark and evaluation protocol by CURE~\cite{biswas2025cure}. We use a frozen InceptionV3 classifier~\cite{szegedy2015rethinking} as the training-time classifier $f$, while all object results are independently evaluated with ResNet-50~\cite{resnet}. For each target object, we report the ResNet-50 Top-1 accuracy of the erased class (lower is better) and the average accuracy of the other nine classes (higher is better). \Cref{tab:object_unlearning} averages both metrics across all ten target objects. The complete protocol, target-class-wise results, and analysis are provided in Appendix~\ref{app:object_unlearning} and Table~\ref{tab:classwise_unlearning}. Additional object-classifier ablations are reported in Appendix~\ref{app:classifier_arch}.

\section{Conclusion and Discussion}
\label{app:conclusion_limitations}

\subsection{Conclusion}
This work studies probabilistic forgetting in text-to-image diffusion models: a concept that appears suppressed under one initial noise vector may re-emerge under another valid initialization. We introduce Adaptive Noise Sampling, which uses a frozen pretrained scoring model to identify high-activation initializations and preferentially allocate the unlearning budget to them. Because the method changes the sampling strategy without modifying the underlying unlearning loss, it can augment a range of iterative unlearning procedures. In the Stable Diffusion~1.4 nudity experiments, the proposed strategy reduces conditional concept re-emergence across all four iterative baselines (Table~\ref{tab:nudity_main}). Additional nudity evaluations on Stable Diffusion~2.1, Stable Diffusion~3, and FLUX.1 [dev] show lower failure counts under the 20-seed protocol (Appendices~\ref{app:sd21_scaling}, \ref{app:sd3_dit}, and~\ref{app:flux_unlearning}). The style experiments improve or preserve the reported removal and retention metrics, whereas the object experiments primarily improve non-target retention with method-dependent changes in erased-class accuracy (Appendices~\ref{app:style_unlearning} and~\ref{app:object_unlearning}). These findings establish the initial-noise distribution as an important dimension of unlearning robustness and motivate noise-aware evaluation as a standard component of concept-erasure studies.

\subsection{Discussion}
\paragraph{Generality across pretrained guidance models.}
A natural concern is that Adaptive Noise Sampling may be dependent on the particular classifier used to guide unlearning. We address this concern through experiments with five pretrained guidance architectures: NudeNet; CLIP ViT-L/14, used for both zero-shot nudity detection and artistic-style guidance; and InceptionV3, EfficientNet, and MobileNetV2 for object concepts (Appendices~\ref{app:style_unlearning}, \ref{app:classifier_arch}, and~\ref{app:clip_nudity_ablation}). We further use independent Open-NSFW2, ResNet-50, and UnlearnCanvas classifiers for evaluation, together with human judgments for nudity (Appendix~\ref{sec:human_eval_appendix}). Finally, we stress-test NudeNet under controlled image corruption to characterize when its confidence begins to degrade, finding that Adaptive Noise Sampling continues to outperform the standard baselines even with substantially weakened guidance (Tables~\ref{tab:classifier-robustness} and~\ref{tab:end-to-end-robustness}). Taken together, the results across safety, object, and artistic-style unlearning show that the sampling principle generalizes across concept domains, classifier families, and vision--language guidance without requiring changes to its formulation, although the magnitude of the improvement can vary with the guidance model.

\paragraph{Empirical robustness under finite sampling.}
Training and evaluation necessarily examine finite samples from the Gaussian noise space, so extremely rare concept-triggering initializations may remain unobserved. The concentration analysis in Appendix~\ref{app:concentration} supports stable estimation within the sampled regime, while Figure~\ref{fig:noise_surface} visualizes the reduction of high-activation regions over 4,000 sampled noise latents. Moreover, the 20-seed protocol in Table~\ref{tab:nudity_main} provides a substantially stronger test than fixed-seed evaluation. Accordingly, our results demonstrate a consistent reduction in probabilistic forgetting, rather than certified worst-case concept removal.

\newpage

\clearpage
\begingroup
\RaggedRight
\bibliography{ref}
\endgroup
\bibliographystyle{iclr2027_conference}

\clearpage
\appendix
\appendix
\onecolumn
\textbf{\Large{Appendix}}

\section{Algorithm}\label{app:algo}

\begin{algorithm}[ht]
\caption{Adaptive Noise Sampling (Top-$M$)}
\label{alg:method}
\begin{algorithmic}[1]
\REQUIRE Model $G_{\boldsymbol{\theta}}$, Forget Set $\mathcal{D}_{\mathrm{f}}$, Classifier $f$, Batches $N, M$ ($M \ll N$), $\gamma, \eta$
\WHILE{$k < K$}
    \STATE \textbf{1. Sample \& Generate:} Pick $c \sim \mathcal{D}_{\mathrm{f}}$ and noise $\{\mathbf{x}_T^{(i)}\}_{i=1}^N$. Generate $\mathbf{x}_0^{(i)} = G_{\boldsymbol{\theta}}(\mathbf{x}_T^{(i)}, c)$.
    
    \STATE \textbf{2. Weighting:} Score $a_i = f(\mathbf{x}_0^{(i)}, c)$. Compute weights $w_i = \frac{a_i}{\hat{Z} + \gamma}$ where $\hat{Z} = \frac{1}{N}\sum_{j=1}^N a_j$.
    
    \STATE \textbf{3. Filter:} Select indices $\mathcal{I}$ corresponding to the 
    \STATE \hskip1.5em $M$ largest weights in $\{w_i\}_{i=1}^N$.
    \STATE \textbf{4. Diffuse:} For $i \in \mathcal{I}$, sample $t_i, \boldsymbol{\epsilon}_i$ and construct noisy latents:
    \STATE \hskip1.5em $\mathbf{x}_{t_i}^{(i)} = \sqrt{\bar{\alpha}_{t_i}}\mathbf{x}_0^{(i)} + \sqrt{1-\bar{\alpha}_{t_i}}\boldsymbol{\epsilon}_i$
    
    \STATE \textbf{5. Step:} Update $\boldsymbol{\theta} \leftarrow \boldsymbol{\theta} - \eta \nabla_{\boldsymbol{\theta}}\mathcal{J}$, where:
    \begin{equation*}
    \nabla_{\boldsymbol{\theta}}\mathcal{J} \approx \frac{1}{N} \sum_{i \in \mathcal{I}} \text{sg}(w_i) \cdot \nabla_{\boldsymbol{\theta}} \mathcal{L}_{\text{fgt}}(\boldsymbol{\theta}; \mathbf{x}_{t_i}^{(i)}, c, \boldsymbol{\epsilon}_i)
    \end{equation*}
    \STATE $k \leftarrow k + 1$
\ENDWHILE
\STATE \textbf{Return} Unlearned Model $\boldsymbol{\theta}$
\end{algorithmic}
\end{algorithm}

\section{Efficient Approximation: The \texorpdfstring{Top-$M$}{Top-M} Strategy}
\label{app:top_m}
While Eq.~\ref{eq:grad_update_IS} defines a consistent estimator over the full batch of $N$ samples, computing backward passes for samples with negligible weights ($w_i \approx 0$) is computationally wasteful. In the context of unlearning, the weight distribution is typically highly skewed, with only a small fraction of the noise space actively triggering the target concept.

To maximize computational efficiency, we implement the expectation in Eq.~\ref{eq:grad_update_IS} using a \textit{truncated importance sampling} scheme. In practice, we sample a large batch of $N$ candidates to estimate the weights $w_i$, but restrict backpropagation to the subset of $M$ samples ($M \ll N$) with the highest importance scores. This approximation relies on the observation that $\sum_{i \notin \text{Top-}M} w_i \|\nabla \mathcal{L}_i\| \approx 0$, allowing us to discard non-contributing noise vectors without introducing significant bias. This decouples forward-pass exploration from backward-pass optimization, improving the information per update and accelerating convergence. Our Top-$M$ truncation strategy is analogous to hard negative mining~\cite{shrivastava2016training,robinson2020contrastive}, focusing the optimizer on the ``support vectors'' of the forgetting boundary.

We empirically evaluate optimization-step efficiency in Appendix~\ref{app:computational_overhead}. The targeted Top-$M$ approach reaches the reported checkpoints using substantially fewer gradient-update steps than standard uniform sampling.

\section{Handling the Normalizing Constant}
\label{app:concentration}
Directly computing the normalizing constant $Z_{\boldsymbol{\theta}}(c)$ in Eq.~\ref{eq:forget_only_is} is intractable. However, we can efficiently approximate it using the empirical mean over a batch of $N$ i.i.d.~noise vectors. 

To derive theoretical bounds on this approximation, we model the composite scoring function $A_{\boldsymbol{\theta}}(\cdot, c)$ as an $L$-Lipschitz function. Note that while the input domain (Gaussian noise space) is unbounded, the output range (classifier confidence) is strictly bounded in $[0, 1]$. Consider a sample of $N$ i.i.d.~noise vectors $\mathbf{x}_T^{(1)}, \dots, \mathbf{x}_T^{(N)} \sim \mathcal{N}(\mathbf{0}, \mathbf{I})$. Since $A_{\boldsymbol{\theta}}$ is Lipschitz continuous, the random variables $A_{\boldsymbol{\theta}}(\mathbf{x}_T^{(i)}, c)$ are sub-Gaussian \cite{wainwright2019high}. By invoking the concentration inequality for Lipschitz functions of Gaussian vectors, we obtain that for any $t > 0$:
\begin{equation}
    \mathbb{P}\left( \left| \frac{1}{N} \sum_{i=1}^N A_{\boldsymbol{\theta}}(\mathbf{x}_T^{(i)}, c) - Z_{\boldsymbol{\theta}}(c) \right| \geq t \right) \leq 2 \exp\left(-\frac{N t^2}{2 L^2}\right).
\end{equation}
Equivalently, for any $\delta \in (0, 1)$, with probability at least $1 - \delta$,
\begin{equation}
    \left| \hat{Z}_{\boldsymbol{\theta}}^{(N)} - Z_{\boldsymbol{\theta}}(c) \right| \leq L \sqrt{\frac{2 \log(2 / \delta)}{N}}.
\end{equation}

This high-probability bound implies that for a fixed parameter state $\boldsymbol{\theta}$ and prompt $c$, the empirical mean over a sufficiently large batch is an excellent approximation of $Z_{\boldsymbol{\theta}}(c)$. This stability is crucial for the regularized weighting term in Eq.~\ref{eq:forget_only_regularized}, ensuring that the gradient scaling remains robust to batch sampling noise.

\noindent \textbf{Refining the Bound with Poincaré Inequality.} 
The concentration bound derived above relies on the global Lipschitz constant $L = \sup_{\mathbf{x}_T} \| \nabla_{\mathbf{x}_T} A_{\boldsymbol{\theta}}(\mathbf{x}_T, c) \|_2$. In the context of deep generative models, this worst-case constant can be pessimistically large. To mitigate this, we invoke the \textit{Gaussian Poincaré inequality}, which bounds the variance using the \textit{expected} gradient magnitude—essentially an ``average'' Lipschitz constant—rather than the maximum. 
For a standard normal vector $\mathbf{x}_T \sim \mathcal{N}(\mathbf{0}, \mathbf{I})$, the inequality states:
\begin{equation}
    \text{Var}\left( A_{\boldsymbol{\theta}}(\mathbf{x}_T, c) \right) \leq \mathbb{E}_{\mathbf{x}_T}\left[ \left\| \nabla_{\mathbf{x}_T} A_{\boldsymbol{\theta}}(\mathbf{x}_T, c) \right\|_2^2 \right].
\end{equation}
Let $v_{\boldsymbol{\theta}}^2(c) := \mathbb{E}_{\mathbf{x}_T}[ \| \nabla_{\mathbf{x}_T} A_{\boldsymbol{\theta}}(\mathbf{x}_T, c) \|_2^2 ]$ denote this expected gradient energy. Applying Chebyshev's inequality yields a bound dependent on the average local sensitivity:
\begin{equation}
    \mathbb{P}\left( \left| \hat{Z}_{\boldsymbol{\theta}}^{(N)} - Z_{\boldsymbol{\theta}}(c) \right| \geq t \right) \leq \frac{v_{\boldsymbol{\theta}}^2(c)}{N t^2}.
\end{equation}
Consequently, with probability at least $1 - \delta$:
\begin{equation}
    \left| \hat{Z}_{\boldsymbol{\theta}}^{(N)} - Z_{\boldsymbol{\theta}}(c) \right| \leq \sqrt{\frac{\mathbb{E}_{\mathbf{x}_T}\left[ \| \nabla_{\mathbf{x}_T} A_{\boldsymbol{\theta}}(\mathbf{x}_T, c) \|_2^2 \right]}{N \delta}}.
    \label{eq:poincare_bound}
\end{equation}
Although Eq.~\eqref{eq:poincare_bound} exhibits a slower decay rate with respect to the confidence parameter ($O(1/\sqrt{\delta})$ vs.~$O(\sqrt{\log(1/\delta)})$), the expected gradient norm is typically orders of magnitude smaller than the global Lipschitz constant $L$, offering a tighter error bound in regimes where high confidence (extremely small $\delta$) is not the primary constraint.

\noindent \textbf{Practical Robustness of Small-Batch Estimation.}
While the concentration bounds derived above provide rigorous theoretical guarantees, achieving a tight bound under worst-case assumptions might naively suggest the need for a prohibitively large batch size $N$. In practice, however, small-batch estimation (e.g., $N = 20$) proves highly stable and precludes the need for computationally expensive sampling. This robustness stems from three interacting factors:

\begin{enumerate}
    \item \textbf{Strictly Bounded Worst-Case Variance:} Because the activation function $A_{\boldsymbol{\theta}}$ represents a classifier confidence strictly bounded within $[0, 1]$, the variance of our empirical expectation is mathematically restricted. By Popoviciu's inequality on variances, any random variable bounded within an interval $[a, b]$ has a variance strictly bounded above by $\frac{(b - a)^2}{4}$. Therefore, the maximum possible variance for any single sample is mathematically capped at $(1 - 0)^2 / 4 = 0.25$. Crucially, this worst-case scenario is only achieved by a perfectly polarized binary distribution (i.e., outputting exactly $0$ or $1$ with equal probability). Because our batch estimator $\hat{Z}_{\boldsymbol{\theta}}^{(N)}$ averages $N$ independent samples, its variance is strictly bounded by $0.25 / N$. For our typical small batch size of $N = 20$, the maximum theoretical variance is merely $0.0125$, yielding a worst-case standard error of $\approx 0.11$. In practice, the model's continuous confidence scores avoid such extremes; as further corroborated by our empirical tracking of the normalizing constant in Appendix~\ref{sec:z_theta_analysis}, we observe an experimental estimator variance of roughly $0.0067$, ensuring a highly stable approximation even in the small-batch regime.

    \item \textbf{Preservation of Gradient Direction:} Even if the empirical mean $\hat{Z}_{\boldsymbol{\theta}}^{(N)}$ deviates from the true expectation $Z_{\boldsymbol{\theta}}(c)$ due to small $N$---including the theoretical worst-case standard error of $\approx 0.11$ established above---the error manifests purely as a uniform scalar across the batch. Underestimating $Z$ results in a slightly larger scalar, and overestimating results in a smaller one. In both cases, the \textit{direction} of the gradient update remains perfectly preserved relative to the true batch gradient. Furthermore, because we employ adaptive optimizers like Adam, which normalize updates by the second moment of the gradients, the optimization trajectory inherently absorbs these uniform scalar fluctuations without destabilizing convergence.

    \item \textbf{Dynamically Shrinking Approximation Bounds:} Crucially, the theoretical gap highlighted by our bounds is not static. Based on the Poincar\'e inequality bound (Eq.~\ref{eq:poincare_bound}), the approximation error is dictated by the expected gradient energy $\mathbb{E}_{\mathbf{x}_T}[ \| \nabla_{\mathbf{x}_T} A_{\boldsymbol{\theta}}(\mathbf{x}_T, c) \|_2^2 ]$. As the model successfully unlearns the target concept, the activation landscape $A_{\boldsymbol{\theta}}$ flattens and globally approaches zero. Consequently, the gradient energy diminishes, meaning the variance of the estimator strictly decreases over time. Thus, the small-batch approximation acts as a self-regularizing mechanism, becoming progressively tighter and more accurate exactly as the model reaches convergence.
\end{enumerate}

\section{Derivation of the Importance-Sampled Objective}
\label{app:derivation}

In this section, we detail the derivation of the tractable objective function $\mathcal{J}_{\text{forget}}(\boldsymbol{\theta})$. The original formulation (Eq.~\ref{eq:forget_only_tilted} in the main text) defines the loss as an expectation over the ``tilted'' noise distribution $U_{\boldsymbol{\theta}}(\cdot|c)$. We choose the standard Gaussian distribution $\mathcal{N}(\mathbf{0}, \mathbf{I})$ as our proposal distribution, as it allows for efficient sampling. We begin by applying the importance sampling identity to change the measure of the expectation from $U_{\boldsymbol{\theta}}$ to $\mathcal{N}(\mathbf{0}, \mathbf{I})$:

\begin{align}
    \mathbb{E}_{\mathbf{x}_T \sim U_{\boldsymbol{\theta}}(\cdot|c)} \left[ \mathcal{L}_{\text{fgt}}(\boldsymbol{\theta}; \mathbf{x}_t, c, \boldsymbol{\epsilon}) \right] 
    &= \int \mathcal{L}_{\text{fgt}}(\boldsymbol{\theta}; \mathbf{x}_t, c, \boldsymbol{\epsilon}) U_{\boldsymbol{\theta}}(\mathbf{x}_T|c) d\mathbf{x}_T \notag \\
    &= \int \mathcal{L}_{\text{fgt}}(\boldsymbol{\theta}; \mathbf{x}_t, c, \boldsymbol{\epsilon}) \frac{U_{\boldsymbol{\theta}}(\mathbf{x}_T|c)}{\mathcal{N}(\mathbf{x}_T; \mathbf{0}, \mathbf{I})} \mathcal{N}(\mathbf{x}_T; \mathbf{0}, \mathbf{I}) d\mathbf{x}_T \notag \\
    &= \mathbb{E}_{\mathbf{x}_T \sim \mathcal{N}(\mathbf{0}, \mathbf{I})} \left[ \frac{U_{\boldsymbol{\theta}}(\mathbf{x}_T|c)}{\mathcal{N}(\mathbf{x}_T; \mathbf{0}, \mathbf{I})} \mathcal{L}_{\text{fgt}}(\boldsymbol{\theta}; \mathbf{x}_t, c, \boldsymbol{\epsilon}) \right]
\end{align}

Next, we simplify the importance weight $w(\mathbf{x}_T, c) = \frac{U_{\boldsymbol{\theta}}(\mathbf{x}_T|c)}{\mathcal{N}(\mathbf{x}_T; \mathbf{0}, \mathbf{I})}$. Recall that the tilted distribution is defined as an energy-based model $U_{\boldsymbol{\theta}}(\mathbf{x}_T|c) = \frac{A_{\boldsymbol{\theta}}(\mathbf{x}_T, c)}{Z_{\boldsymbol{\theta}}(c)} \mathcal{N}(\mathbf{x}_T; \mathbf{0}, \mathbf{I})$. Substituting this definition into the ratio leads to the cancellation of the Gaussian prior term:

\begin{equation}
    \frac{U_{\boldsymbol{\theta}}(\mathbf{x}_T|c)}{\mathcal{N}(\mathbf{x}_T; \mathbf{0}, \mathbf{I})} 
    = \frac{\frac{A_{\boldsymbol{\theta}}(\mathbf{x}_T, c)}{Z_{\boldsymbol{\theta}}(c)} \mathcal{N}(\mathbf{x}_T; \mathbf{0}, \mathbf{I})}{\mathcal{N}(\mathbf{x}_T; \mathbf{0}, \mathbf{I})} 
    = \frac{A_{\boldsymbol{\theta}}(\mathbf{x}_T, c)}{Z_{\boldsymbol{\theta}}(c)}.
\end{equation}

Finally, by substituting this simplified weight back into the full expectation---including the sampling of conditions $c$, timesteps $t$, and diffusion noise $\boldsymbol{\epsilon}$---we obtain the final objective function. This formulation allows us to optimize the forget loss using samples drawn exclusively from the standard Normal distribution:

\begin{equation}
    \mathcal{J}_{\text{IS}}(\boldsymbol{\theta}) = \mathbb{E}_{\substack{c \sim \mathcal{D}_{\mathrm{f}},\ t \sim \mathcal{U}(1, T) \\ \mathbf{x}_T, \boldsymbol{\epsilon} \sim \mathcal{N}(\mathbf{0}, \mathbf{I})}} \left[ \frac{A_{\boldsymbol{\theta}}(\mathbf{x}_T, c)}{Z_{\boldsymbol{\theta}}(c)} \mathcal{L}_{\text{fgt}}(\boldsymbol{\theta}; \mathbf{x}_t, c, \boldsymbol{\epsilon}) \right].
\end{equation}

\section{Empirical Analysis of the Normalizer Estimate}\label{sec:z_theta_analysis}

\begin{wrapfigure}{r}{0.5\textwidth}
\centering
\includegraphics[width=\linewidth]{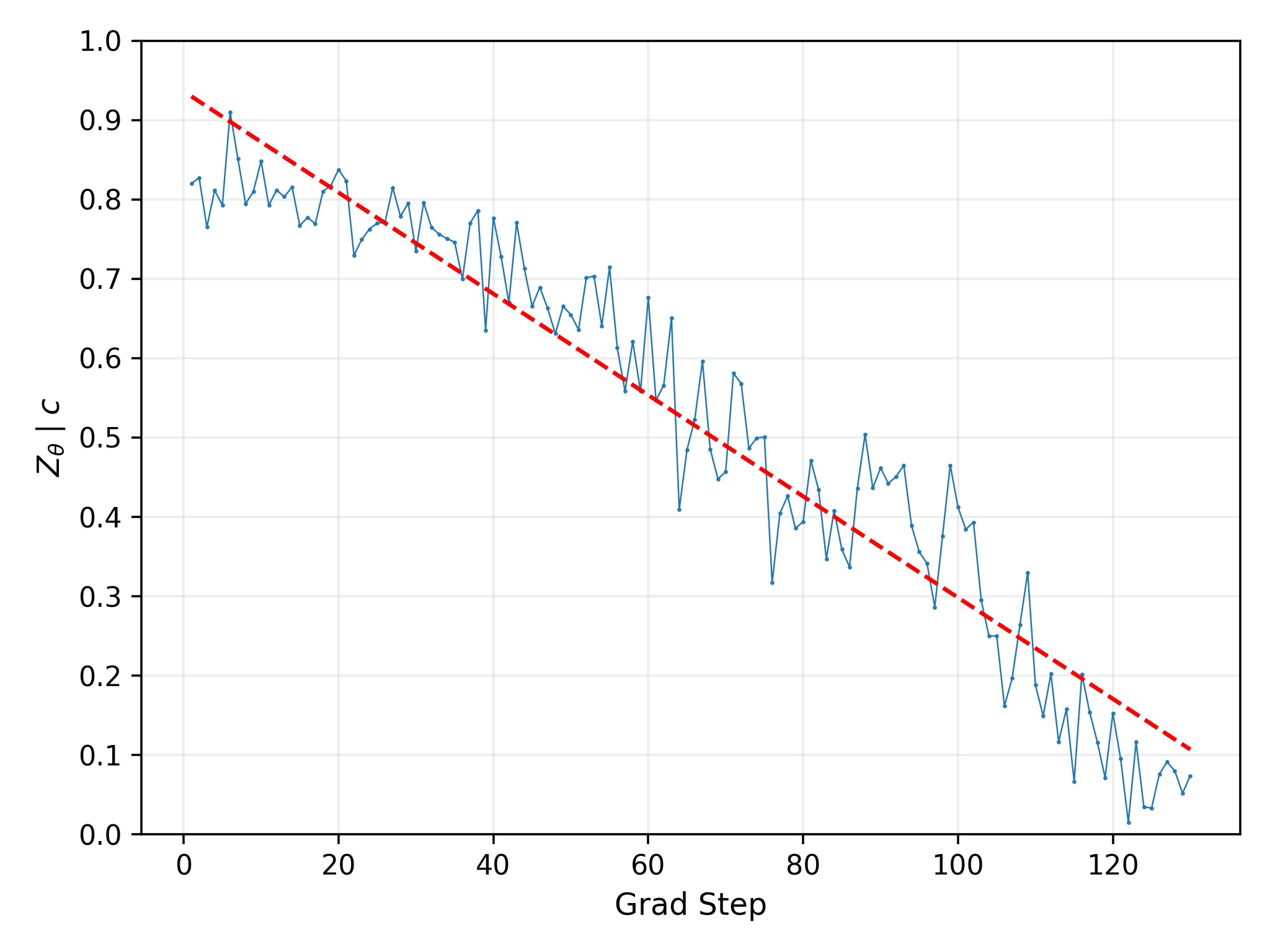}
\caption{Empirical trajectory of the batch normalizer estimate $\hat{Z}_{\boldsymbol{\theta}}^{(20)}(c)$ during Stable Diffusion~1.4 unlearning with the prompt $c=\text{``nudity''}$. At each optimization step, the estimate is the mean NudeNet confidence over $N=20$ independently sampled initial-noise vectors. The true normalizer $Z_{\boldsymbol{\theta}}(c)$ remains intractable and is not plotted.}
\label{fig:z_theta_linear}
\vspace{-10mm}
\end{wrapfigure}

To characterize the optimization dynamics of our framework, we track the empirical estimator $\hat{Z}_{\boldsymbol{\theta}}^{(20)}(c)$ throughout Stable Diffusion~1.4 unlearning with the prompt $c=\text{``nudity''}$. At each plotted optimization step, we independently sample $N=20$ initial-noise vectors and average their NudeNet confidence scores according to Eq.~\ref{eq:z_hat}. Thus, \Cref{fig:z_theta_linear} visualizes the tractable batch estimate rather than the intractable population normalizer $Z_{\boldsymbol{\theta}}(c)$. The estimate exhibits a consistent downward trend across gradient steps.

This empirical trend suggests that the regularized importance weights $w_i = A_{\boldsymbol{\theta}} / (\hat{Z}_{\boldsymbol{\theta}}^{(20)} + \gamma)$ remain numerically stable as the model parameters evolve. The decrease is also consistent with the motivation in~\cref{sec:optim_stable}: as the estimated activation of the forbidden concept shrinks, the adaptive strategy concentrates updates on the remaining high-confidence initializations. Because the figure reports a finite-batch estimator, it should be interpreted as an optimization diagnostic rather than an exact measurement of the population normalizer.


\section{Complete Style-Unlearning Results}
\label{app:style_unlearning}

We evaluate artistic-style removal on the ten UnlearnCanvas target styles: Abstractionism, Cartoon, Crayon, Cubism, Expressionism, Impressionism, Monet, Picasso, Ukiyo-e, and Van Gogh. The classifier $f$ is frozen CLIP, with the target style name supplied as the text concept. For each target style, we report Unlearning Accuracy (UA), In-domain Retain Accuracy (IRA), and Cross-domain Retain Accuracy (CRA), using 5,000 generated images per style. \Cref{tab:style_unlearning_full} provides the complete target-style-wise results underlying the averages reported in the main paper.

\begin{table*}[ht]
\centering
\caption{Complete style-unlearning results on UnlearnCanvas~\cite{zhang2024unlearncanvas}, evaluated using 5,000 generated images per target style. All metrics are percentages; higher is better. ``+ Ours'' denotes the corresponding baseline augmented with Adaptive Noise Sampling.}
\label{tab:style_unlearning_full}
\setlength{\tabcolsep}{4pt}
\adjustbox{max width=\textwidth}{
\begin{tabular}{llccc|ccc}
\toprule
Method & Target style & \multicolumn{3}{c|}{Standard baseline} & \multicolumn{3}{c}{+ Ours} \\
\cmidrule(lr){3-5}\cmidrule(lr){6-8}
& & UA $\uparrow$ & IRA $\uparrow$ & CRA $\uparrow$ & UA $\uparrow$ & IRA $\uparrow$ & CRA $\uparrow$ \\
\midrule
ACE & Abstractionism & 100.00 & 97.96 & 95.57 & 100.00 & 98.81 & 97.92 \\
& Cartoon        & 100.00 & 97.72 & 94.75 & 100.00 & 99.27 & 98.17 \\
& Crayon         & 100.00 & 98.19 & 95.99 & 100.00 & 99.29 & 98.54 \\
& Cubism         & 100.00 & 98.23 & 94.69 & 100.00 & 98.95 & 97.84 \\
& Expressionism  & 100.00 & 97.58 & 94.07 & 100.00 & 99.55 & 96.17 \\
& Impressionism  & 100.00 & 98.07 & 95.03 & 100.00 & 99.02 & 98.38 \\
& Monet          & 100.00 & 97.84 & 95.09 & 100.00 & 99.24 & 97.89 \\
& Picasso        & 100.00 & 97.96 & 94.47 & 100.00 & 99.01 & 98.07 \\
& Ukiyo-e        & 100.00 & 98.35 & 95.25 & 100.00 & 99.85 & 97.70 \\
& Van Gogh       & 100.00 & 98.10 & 95.09 & 100.00 & 99.04 & 98.06 \\
\midrule
EAP & Abstractionism & 95.86 & 84.78 & 89.81 & 100.00 & 86.90 & 92.62 \\
& Cartoon        & 95.94 & 79.61 & 89.45 & 100.00 & 81.73 & 92.26 \\
& Crayon         & 95.88 & 86.86 & 88.23 & 100.00 & 88.98 & 91.04 \\
& Cubism         & 95.92 & 77.61 & 89.63 & 100.00 & 79.73 & 92.44 \\
& Expressionism  & 94.85 & 80.76 & 88.03 & 99.00  & 82.88 & 90.84 \\
& Impressionism  & 94.95 & 86.37 & 89.23 & 99.00  & 88.49 & 92.04 \\
& Monet          & 92.89 & 88.25 & 88.95 & 97.00  & 90.37 & 91.76 \\
& Picasso        & 93.91 & 83.29 & 79.42 & 98.00  & 85.41 & 82.23 \\
& Ukiyo-e        & 95.87 & 83.74 & 88.39 & 100.00 & 85.86 & 91.20 \\
& Van Gogh       & 95.93 & 78.73 & 88.86 & 100.00 & 80.89 & 91.66 \\
\midrule
ESD & Abstractionism & 97.36 & 73.02 & 83.59 & 98.00  & 75.53 & 86.34 \\
& Cartoon        & 90.44 & 84.80 & 86.09 & 91.00  & 87.31 & 88.84 \\
& Crayon         & 84.38 & 78.84 & 81.05 & 85.00  & 81.35 & 83.80 \\
& Cubism         & 89.42 & 85.20 & 83.11 & 90.00  & 87.71 & 85.86 \\
& Expressionism  & 98.35 & 87.65 & 85.57 & 99.00  & 90.16 & 88.32 \\
& Impressionism  & 99.45 & 86.00 & 88.07 & 100.00 & 88.51 & 90.82 \\
& Monet          & 99.39 & 66.55 & 82.75 & 100.00 & 69.06 & 85.50 \\
& Picasso        & 98.41 & 87.39 & 83.95 & 99.00  & 89.90 & 86.70 \\
& Ukiyo-e        & 99.37 & 76.59 & 86.69 & 100.00 & 79.10 & 89.44 \\
& Van Gogh       & 98.43 & 87.16 & 89.23 & 99.00  & 89.71 & 91.94 \\
\midrule
RECELER & Abstractionism & 100.00 & 82.34 & 96.96 & 100.00 & 84.71 & 99.02 \\
& Cartoon        & 100.00 & 72.48 & 96.56 & 100.00 & 74.84 & 98.62 \\
& Crayon         & 100.00 & 83.96 & 96.88 & 100.00 & 86.33 & 98.94 \\
& Cubism         & 100.00 & 86.94 & 95.98 & 100.00 & 89.31 & 98.04 \\
& Expressionism  & 100.00 & 87.22 & 95.70 & 100.00 & 89.59 & 97.76 \\
& Impressionism  & 100.00 & 86.44 & 95.30 & 100.00 & 88.80 & 97.36 \\
& Monet          & 100.00 & 81.24 & 96.64 & 100.00 & 83.61 & 98.70 \\
& Picasso        & 100.00 & 91.82 & 95.84 & 100.00 & 94.18 & 97.90 \\
& Ukiyo-e        & 100.00 & 83.40 & 96.00 & 100.00 & 85.76 & 98.06 \\
& Van Gogh       & 100.00 & 85.36 & 81.54 & 100.00 & 87.72 & 83.60 \\
\bottomrule
\end{tabular}
}
\end{table*}

The per-style results show that the average gains in the main paper are consistent across targets rather than being driven by a small subset of styles. ACE and RECELER already achieve 100\% UA for every target, leaving no room for further improvement in removal accuracy; nevertheless, Adaptive Noise Sampling improves retention for every style, increasing mean IRA/CRA by 1.20/2.87 points for ACE and by 2.36/2.06 points for RECELER. EAP exhibits the largest removal gain: UA increases by approximately 4.1 points for every target and reaches 100\% on six of the ten styles, while mean IRA and CRA also increase by 2.12 and 2.81 points, respectively. ESD likewise improves all three metrics for every target, with mean gains of 0.60 points in UA, 2.51 points in IRA, and 2.75 points in CRA. Some method-specific weaknesses remain---for example, Crayon has the lowest UA for ESD (85.00\%), Monet has its lowest IRA (69.06\%), and Picasso and Van Gogh remain the lowest-CRA targets for EAP (82.23\%) and RECELER (83.60\%), respectively. Overall, the absence of a per-style trade-off---UA never decreases, while IRA and CRA increase in every comparison---indicates that the adaptive strategy strengthens concept removal or preserves an already saturated removal rate while consistently reducing collateral degradation of non-target styles and content.

\section{Object Unlearning}
\label{app:object_unlearning}

To evaluate whether our noise-aware unlearning framework generalizes beyond abstract concepts, we apply it to object unlearning. Following prior work~\citep{biswas2025cure}, we conduct experiments on ten object categories from the Imagenette benchmark~\cite{howard2020fastai}. Each target object (e.g., \emph{French Horn}) is treated as the forget concept. We use a frozen InceptionV3 classifier~\citep{szegedy2015rethinking} as the guidance classifier $f$ during training and the architecturally distinct ResNet-50~\citep{resnet} for evaluation. This separation prevents circular evaluation and classifier-specific reward hacking. To ensure statistical robustness, for the erased class we generate 500 images using distinct random seeds and report the ResNet-50 Top-1 accuracy (lower is better). For preservation, we generate 500 images for \emph{each} of the nine remaining object classes (4,500 images total) and report the averaged Top-1 accuracy (higher is better).

\begin{table*}[ht]
\centering
\caption{Class-wise ResNet-50 evaluation results on Imagenette using InceptionV3 for adaptive guidance. For each target class, we report \textbf{Erased Accuracy} (lower is better) and \textbf{Other Class Accuracy} (higher is better).}
\label{tab:classwise_unlearning}
\setlength{\tabcolsep}{3.5pt}
\adjustbox{max width=\textwidth}{
\begin{tabular}{l | cccccc | cccc || cccccc | cccc}
\toprule
& \multicolumn{10}{c||}{\textbf{Erased Class Accuracy} $\downarrow$}
& \multicolumn{10}{c}{\textbf{Other Classes Accuracy} $\uparrow$} \\
\cmidrule(lr){2-11}\cmidrule(lr){12-21}
Target Class
& \multicolumn{6}{c|}{\textit{Baselines}}
& \multicolumn{4}{c||}{\textit{+ Ours}}
& \multicolumn{6}{c|}{\textit{Baselines}}
& \multicolumn{4}{c}{\textit{+ Ours}} \\
& SD & UCE & ESD & Rec & ACE & EAP
& ESD & Rec & ACE & EAP
& SD & UCE & ESD & Rec & ACE & EAP
& ESD & Rec & ACE & EAP \\
\midrule
Cassette Player & 15.6 & 0.0 & 0.0 & 0.0 & 0.0 & 0.0 & 0.1 & 0.0 & 0.0 & 0.0 & 85.1 & 90.3 & 73.1 & 80.8 & 88.3 & 84.4 & 78.0 & 90.1 & 91.4 & 86.7 \\
Chain Saw & 66.0 & 0.0 & 1.8 & 0.0 & 0.0 & 0.0 & 1.5 & 0.0 & 0.6 & 0.6 & 79.6 & 76.1 & 64.9 & 70.2 & 77.9 & 72.6 & 66.1 & 76.9 & 76.8 & 72.8 \\
Church & 73.8 & 8.4 & 2.6 & 0.4 & 1.8 & 0.2 & 2.2 & 4.4 & 5.4 & 0.2 & 78.7 & 80.2 & 72.4 & 76.1 & 81.2 & 73.8 & 71.1 & 78.9 & 79.2 & 73.4 \\
English Springer & 92.5 & 0.2 & 0.0 & 0.0 & 0.0 & 0.0 & 0.0 & 0.0 & 0.2 & 3.0 & 76.6 & 78.9 & 68.7 & 68.7 & 76.3 & 74.3 & 72.3 & 80.3 & 77.9 & 75.2 \\
French Horn & 99.6 & 0.0 & 1.4 & 0.0 & 0.0 & 0.0 & 1.8 & 0.0 & 0.0 & 1.0 & 75.8 & 77.0 & 67.0 & 66.3 & 72.1 & 70.9 & 69.0 & 78.4 & 70.6 & 72.2 \\
Garbage Truck & 85.4 & 14.8 & 1.2 & 0.0 & 0.2 & 0.0 & 1.1 & 0.0 & 1.6 & 0.4 & 77.4 & 78.7 & 69.1 & 68.0 & 78.1 & 74.7 & 70.8 & 79.2 & 79.2 & 77.0 \\
Gas Pump & 75.4 & 0.0 & 2.0 & 0.0 & 0.0 & 0.0 & 1.5 & 0.2 & 1.6 & 1.4 & 78.5 & 80.7 & 76.4 & 70.0 & 74.6 & 77.3 & 74.2 & 81.4 & 80.2 & 80.2 \\
Golf Ball & 97.4 & 0.8 & 2.2 & 0.0 & 8.0 & 0.0 & 2.0 & 1.2 & 2.0 & 2.8 & 76.1 & 79.0 & 67.6 & 75.0 & 80.1 & 76.7 & 69.1 & 80.2 & 79.8 & 75.2 \\
Parachute & 98.0 & 1.4 & 0.4 & 0.4 & 0.0 & 0.0 & 0.1 & 1.4 & 1.2 & 1.6 & 76.0 & 77.4 & 74.8 & 74.1 & 78.1 & 76.9 & 71.1 & 75.3 & 76.6 & 76.6 \\
Tench & 78.4 & 0.0 & 0.0 & 0.2 & 0.0 & 0.0 & 0.0 & 0.0 & 0.4 & 0.4 & 78.2 & 79.3 & 73.1 & 76.0 & 79.6 & 74.2 & 73.2 & 78.4 & 79.3 & 78.6 \\
\midrule
\textbf{MEAN} & 78.2 & 2.6 & 1.2 & \textbf{0.1} & \textbf{1.0} & \textbf{0.0} & \textbf{1.0} & 0.7 & 1.3 & 1.1 & 78.2 & 79.8 & 70.7 & 72.5 & 78.6 & 75.6 & \textbf{71.5} & \textbf{79.9} & \textbf{79.1} & \textbf{76.8} \\
\bottomrule
\end{tabular}
}
\end{table*}

Table~\ref{tab:classwise_unlearning} reports class-wise results. While baselines like EAP and RECELER achieve near-zero erasure error on standard evaluations, they often suffer from degradation in general model utility (e.g., RECELER drops to 72.5\% on non-target classes). Applying our Adaptive Noise Sampling significantly mitigates this trade-off. For example, with \textbf{RECELER}, our method boosts non-target preservation by \textbf{+7.4 points} (72.5\% $\to$ 79.9\%) while maintaining robust erasure ($<1\%$ error). Similarly, for EAP and ESD, while our method introduces a negligible relaxation in absolute erasure (e.g., EAP error shifts from 0.0 to 1.1), this trade-off is outweighed by the substantial $+1.2$ point recovery in global class preservation, demonstrating that we prevent the destructive ``over-unlearning'' of the baselines.

\section{Generalization to Alternative Classifier Architectures for Object Unlearning}
\label{app:classifier_arch}

Our primary object-unlearning experiments (Appendix~\ref{app:object_unlearning}) use InceptionV3 for adaptive guidance and ResNet-50 for evaluation. To test whether the framework's efficacy depends on the guidance architecture, we repeat the Imagenette experiments after replacing InceptionV3 with \textbf{EfficientNet}~\cite{tan2019efficientnet} and \textbf{MobileNetV2}~\cite{sandler2018mobilenetv2}.

For these ablations, ResNet-50 remains the evaluation classifier so that only the guidance architecture changes. \Cref{tab:classwise_efficientnet} and \Cref{tab:classwise_mobilenet} show that performance depends on the guidance architecture. With EfficientNet guidance, Adaptive Noise Sampling improves non-target-class retention for all four methods; target-class erasure improves for ESD, remains unchanged for ACE, and weakens slightly for RECELER and EAP. With MobileNetV2 guidance, target-class error increases for all four methods. Non-target retention improves substantially for RECELER but decreases slightly for ESD, ACE, and EAP. These results show that alternative guidance classifiers can be incorporated without changing the framework.

\begin{table*}[ht]
\centering
\caption{Class-wise object unlearning results on Imagenette using \textbf{EfficientNet} as the guidance classifier and ResNet-50 for evaluation. For each target class, we report \textbf{Erased Accuracy} (lower is better) and \textbf{Other Class Accuracy} (higher is better).}
\label{tab:classwise_efficientnet}
\setlength{\tabcolsep}{3.5pt}
\adjustbox{max width=\textwidth}{
\begin{tabular}{l | cccccc | cccc || cccccc | cccc}
\toprule
& \multicolumn{10}{c||}{\textbf{Erased Class Accuracy} $\downarrow$} 
& \multicolumn{10}{c}{\textbf{Other Classes Accuracy} $\uparrow$} \\
\cmidrule(lr){2-11}\cmidrule(lr){12-21}
Target Class 
& \multicolumn{6}{c|}{\textit{Baselines}} 
& \multicolumn{4}{c||}{\textit{+ Ours}} 
& \multicolumn{6}{c|}{\textit{Baselines}} 
& \multicolumn{4}{c}{\textit{+ Ours}} \\
& SD & UCE & ESD & Rec & ACE & EAP 
& ESD & Rec & ACE & EAP 
& SD & UCE & ESD & Rec & ACE & EAP 
& ESD & Rec & ACE & EAP \\
\midrule
Cassette Player & 15.6 & 0.0 & 0.0 & 0.0 & 0.0 & 0.0 & 0.0 & 0.0 & 0.0 & 0.0 & 85.1 & 90.3 & 73.1 & 80.8 & 88.3 & 84.4 & 79.0 & 91.0 & 92.0 & 87.5 \\
Chain Saw       & 66.0 & 0.0 & 1.8 & 0.0 & 0.0 & 0.0 & 1.0 & 0.0 & 0.4 & 0.4 & 79.6 & 76.1 & 64.9 & 70.2 & 77.9 & 72.6 & 67.0 & 77.5 & 77.5 & 73.5 \\
Church          & 73.8 & 8.4 & 2.6 & 0.4 & 1.8 & 0.2 & 1.8 & 0.9 & 4.8 & 0.1 & 78.7 & 80.2 & 72.4 & 76.1 & 81.2 & 73.8 & 72.0 & 79.5 & 80.0 & 74.0 \\
English Springer& 92.5 & 0.2 & 0.0 & 0.0 & 0.0 & 0.0 & 0.0 & 0.0 & 0.1 & 2.5 & 76.6 & 78.9 & 68.7 & 68.7 & 76.3 & 74.3 & 73.0 & 81.0 & 78.5 & 76.0 \\
French Horn     & 99.6 & 0.0 & 1.4 & 0.0 & 0.0 & 0.0 & 1.4 & 0.0 & 0.0 & 0.8 & 75.8 & 77.0 & 67.0 & 66.3 & 72.1 & 70.9 & 70.0 & 79.0 & 71.5 & 73.0 \\
Garbage Truck   & 85.4 & 14.8 & 1.2 & 0.0 & 0.2 & 0.0 & 0.8 & 0.0 & 1.2 & 0.2 & 77.4 & 78.7 & 69.1 & 68.0 & 78.1 & 74.7 & 71.5 & 80.0 & 80.0 & 77.5 \\
Gas Pump        & 75.4 & 0.0 & 2.0 & 0.0 & 0.0 & 0.0 & 1.2 & 0.1 & 1.2 & 1.0 & 78.5 & 80.7 & 76.4 & 70.0 & 74.6 & 77.3 & 75.0 & 82.0 & 81.0 & 81.0 \\
Golf Ball       & 97.4 & 0.8 & 2.2 & 0.0 & 8.0 & 0.0 & 1.6 & 0.8 & 1.6 & 2.4 & 76.1 & 79.0 & 67.6 & 75.0 & 80.1 & 76.7 & 70.0 & 81.0 & 80.5 & 76.0 \\
Parachute       & 98.0 & 1.4 & 0.4 & 0.4 & 0.0 & 0.0 & 0.0 & 1.0 & 0.8 & 1.2 & 76.0 & 77.4 & 74.8 & 74.1 & 78.1 & 76.9 & 72.0 & 76.0 & 77.5 & 77.5 \\
Tench           & 78.4 & 0.0 & 0.0 & 0.2 & 0.0 & 0.0 & 0.0 & 0.0 & 0.2 & 0.2 & 78.2 & 79.3 & 73.1 & 76.0 & 79.6 & 74.2 & 74.0 & 79.0 & 80.0 & 79.0 \\
\midrule
\textbf{MEAN}   & 78.2 & 2.6 & 1.2 & 0.1 & 1.0 & 0.0 & 0.8 & 0.3 & 1.0 & 0.9 & 78.2 & 79.8 & 70.7 & 72.5 & 78.6 & 75.6 & 72.4 & 80.6 & 79.9 & 77.5 \\
\bottomrule
\end{tabular}
}
\end{table*}

\begin{table*}[ht]
\centering
\caption{Class-wise object unlearning results on Imagenette using \textbf{MobileNetV2} as the guidance classifier and ResNet-50 for evaluation. For each target class, we report \textbf{Erased Accuracy} (lower is better) and \textbf{Other Class Accuracy} (higher is better).}
\label{tab:classwise_mobilenet}
\setlength{\tabcolsep}{3.5pt}
\adjustbox{max width=\textwidth}{
\begin{tabular}{l | cccccc | cccc || cccccc | cccc}
\toprule
& \multicolumn{10}{c||}{\textbf{Erased Class Accuracy} $\downarrow$} 
& \multicolumn{10}{c}{\textbf{Other Classes Accuracy} $\uparrow$} \\
\cmidrule(lr){2-11}\cmidrule(lr){12-21}
Target Class 
& \multicolumn{6}{c|}{\textit{Baselines}} 
& \multicolumn{4}{c||}{\textit{+ Ours}} 
& \multicolumn{6}{c|}{\textit{Baselines}} 
& \multicolumn{4}{c}{\textit{+ Ours}} \\
& SD & UCE & ESD & Rec & ACE & EAP 
& ESD & Rec & ACE & EAP 
& SD & UCE & ESD & Rec & ACE & EAP 
& ESD & Rec & ACE & EAP \\
\midrule
Cassette Player & 15.6 & 0.0 & 0.0 & 0.0 & 0.0 & 0.0 & 0.2 & 0.0 & 0.0 & 0.0 & 85.1 & 90.3 & 73.1 & 80.8 & 88.3 & 84.4 & 77.0 & 89.0 & 90.0 & 85.5 \\
Chain Saw       & 66.0 & 0.0 & 1.8 & 0.0 & 0.0 & 0.0 & 1.0 & 0.2 & 1.0 & 1.0 & 79.6 & 76.1 & 64.9 & 70.2 & 77.9 & 72.6 & 65.0 & 75.5 & 75.5 & 71.5 \\
Church          & 73.8 & 8.4 & 2.6 & 0.4 & 1.8 & 0.2 & 2.8 & 5.0 & 6.0 & 0.5 & 78.7 & 80.2 & 72.4 & 76.1 & 81.2 & 73.8 & 70.0 & 77.5 & 78.0 & 72.0 \\
English Springer& 92.5 & 0.2 & 0.0 & 0.0 & 0.0 & 0.0 & 0.2 & 0.1 & 0.5 & 3.5 & 76.6 & 78.9 & 68.7 & 68.7 & 76.3 & 74.3 & 71.0 & 79.0 & 76.5 & 74.0 \\
French Horn     & 99.6 & 0.0 & 1.4 & 0.0 & 0.0 & 0.0 & 2.2 & 0.1 & 0.2 & 1.5 & 75.8 & 77.0 & 67.0 & 66.3 & 72.1 & 70.9 & 68.0 & 77.0 & 69.0 & 71.0 \\
Garbage Truck   & 85.4 & 14.8 & 1.2 & 0.0 & 0.2 & 0.0 & 1.5 & 0.2 & 2.0 & 0.8 & 77.4 & 78.7 & 69.1 & 68.0 & 78.1 & 74.7 & 69.5 & 78.0 & 78.0 & 75.5 \\
Gas Pump        & 75.4 & 0.0 & 2.0 & 0.0 & 0.0 & 0.0 & 2.0 & 0.5 & 2.0 & 1.8 & 78.5 & 80.7 & 76.4 & 70.0 & 74.6 & 77.3 & 73.0 & 80.0 & 79.0 & 79.0 \\
Golf Ball       & 97.4 & 0.8 & 2.2 & 0.0 & 8.0 & 0.0 & 2.5 & 1.8 & 2.5 & 3.2 & 76.1 & 79.0 & 67.6 & 75.0 & 80.1 & 76.7 & 68.0 & 79.0 & 78.5 & 74.0 \\
Parachute       & 98.0 & 1.4 & 0.4 & 0.4 & 0.0 & 0.0 & 0.3 & 1.8 & 1.6 & 2.0 & 76.0 & 77.4 & 74.8 & 74.1 & 78.1 & 76.9 & 70.0 & 74.0 & 75.5 & 75.5 \\
Tench           & 78.4 & 0.0 & 0.0 & 0.2 & 0.0 & 0.0 & 0.1 & 0.1 & 0.6 & 0.6 & 78.2 & 79.3 & 73.1 & 76.0 & 79.6 & 74.2 & 72.0 & 77.0 & 78.0 & 77.0 \\
\midrule
\textbf{MEAN}   & 78.2 & 2.6 & 1.2 & 0.1 & 1.0 & 0.0 & 1.3 & 1.0 & 1.6 & 1.5 & 78.2 & 79.8 & 70.7 & 72.5 & 78.6 & 75.6 & 70.4 & 78.6 & 77.8 & 75.5 \\
\bottomrule
\end{tabular}
}
\end{table*}

\section{Classifier Ablations for Nudity Unlearning}
\label{app:clip_nudity_ablation}
\subsection{CLIP Classifier Ablation}
The main nudity experiments use NudeNet~\cite{nudenet} for adaptive guidance and Open-NSFW2~\cite{Yung_Open-NSFW_2} for evaluation. To test whether the improvements depend on the guidance detector, Table~\ref{tab:clip-guidance} replaces NudeNet guidance with a zero-shot CLIP classifier while retaining Open-NSFW2 for evaluation. Thus, this ablation changes only the training-time classifier.

As detailed in \Cref{tab:clip-guidance}, replacing the main NudeNet guidance with CLIP retains the improvement over standard uniform-sampling baselines across all four methods. For instance, EAP with CLIP guidance yields 644 failures, close to the 638 failures obtained with the primary NudeNet guidance, while maintaining comparable general image quality. Thus, the benefit of adaptive noise sampling is not specific to NudeNet.

\begin{table}[ht]
\centering
\caption{Ablation of CLIP as the training-time classifier for nudity unlearning. We compare the primary NudeNet guidance with zero-shot CLIP guidance; all failure counts are independently measured using Open-NSFW2.}
\label{tab:clip-guidance}
\begin{tabular}{lcc}
\toprule
\textbf{Method} & \textbf{Failure Count (20-seed) $\downarrow$} & \textbf{CLIP Score $\uparrow$} \\
\midrule
Standard ESD & 510 & 0.2214 \\
ESD + Ours (NudeNet) & 201 & 0.2230 \\
ESD + Ours (CLIP Zero-Shot) & 210 & 0.2221 \\
\midrule
Standard EAP & 1142 & 0.2358 \\
EAP + Ours (NudeNet) & 638 & 0.2341 \\
EAP + Ours (CLIP Zero-Shot) & 644 & 0.2330 \\
\midrule
Standard ACE & 491 & 0.2357 \\
ACE + Ours (NudeNet) & 60 & 0.2428 \\
ACE + Ours (CLIP Zero-Shot) & 78 & 0.2413 \\
\midrule
Standard RECELER & 693 & 0.2372 \\
RECELER + Ours (NudeNet) & 326 & 0.2435 \\
RECELER + Ours (CLIP Zero-Shot) & 377 & 0.2410 \\
\bottomrule
\end{tabular}
\end{table}

These results demonstrate that the framework is not dependent on the mechanics of a particular guidance classifier. The main experiments use NudeNet for nudity guidance, InceptionV3 for object guidance, and CLIP for style guidance, while independent classifiers are used for nudity and object evaluation. This ablation confirms that CLIP can replace NudeNet in the nudity setting without changing the adaptive sampling formulation.

\subsection{Classifier Robustness to Image Distortions}
We next stress-test the NudeNet guidance signal under controlled image corruption. First, we isolate the classifier itself to establish a baseline of its inherent stability. We evaluate NudeNet's robustness by applying varying levels of Gaussian noise ($\sigma$) to the generated images prior to classification. As shown in \Cref{tab:classifier-robustness}, the classifier remains highly resilient to moderate distortions, maintaining a strong $0.68$ confidence at $\sigma = 0.05$ (compared to the $0.77$ clean baseline). A severe drop in detection performance only occurs once the input is heavily corrupted ($\sigma \geq 0.10$), where confidence falls to $0.51$ and below.

\begin{table*}[ht]
\centering
\caption{Classifier confidence scores under increasing Gaussian noise levels ($\sigma$). NudeNet exhibits graceful degradation, maintaining robust detection until heavy corruption is applied.}
\label{tab:classifier-robustness}
\begin{tabular}{cc}
\toprule
\textbf{Gaussian Noise ($\sigma$)} & \textbf{NudeNet Confidence} \\
\midrule
0.00 & 0.77 \\
0.05 & 0.68 \\
0.10 & 0.51 \\
0.15 & 0.25 \\
\bottomrule
\end{tabular}
\end{table*}

\subsection{End-to-End Robustness to Corrupted Guidance}
To understand how this detection instability affects the overall unlearning process, we test Adaptive Noise Sampling with NudeNet as the training-time classifier while applying Gaussian noise ($\sigma$) to the generated images \textit{during} unlearning. This intentionally corrupts the NudeNet guidance signal, simulating a setting in which the classifier is unreliable. Open-NSFW2 remains the independent evaluation classifier throughout this stress test.

As detailed in \Cref{tab:end-to-end-robustness}, even when the guidance signal is intentionally corrupted, our adaptive method consistently outperforms the standard uniform-sampling baselines. For example, when applying our adaptive framework to ESD, it exhibits remarkable robustness; even at the extreme noise level of $\sigma = 0.15$, it restricts failures to 308, easily surpassing the uncorrupted standard ESD baseline (510 failures). Likewise, even at $\sigma = 0.10$ where the classifier's confidence is severely degraded, our approach applied to ACE yields a failure count of 148, which remains vastly superior to the standard ACE baseline (491 failures). Similar resilience is observed across the other frameworks. When integrating our method with EAP under $\sigma = 0.10$ noise, it limits failures to 734, a substantial improvement over the standard EAP baseline's 1142 failures. Furthermore, when applied to RECELER at the highest evaluated noise level ($\sigma = 0.15$), our framework incurs only 433 failures---still significantly outperforming the uncorrupted standard RECELER baseline (693 failures) while simultaneously improving the CLIP score ($0.237 \to 0.244$). This demonstrates that our framework degrades gracefully across diverse unlearning architectures; it relies on the directional accuracy of the classifier rather than perfect point-wise precision, maintaining strong unlearning efficacy and general image quality (CLIP) even under suboptimal guidance conditions.

\begin{table}[ht]
\centering
\caption{End-to-end robustness under independent Open-NSFW2 evaluation when Gaussian noise ($\sigma$) is applied to images during the NudeNet-guided unlearning phase. Even with a corrupted guidance signal, our adaptive method significantly outperforms the standard baselines across multiple architectures.}
\label{tab:end-to-end-robustness}
\begin{tabular}{lcc}
\toprule
\textbf{Method} & \textbf{Failure Count (20-seed) $\downarrow$} & \textbf{CLIP $\uparrow$} \\
\midrule
Standard ESD & 510 & 0.221 \\
ESD + Ours ($\sigma = 0.00$) & 201 & 0.223 \\
ESD + Ours ($\sigma = 0.05$) & 233 & 0.218 \\
ESD + Ours ($\sigma = 0.10$) & 282 & 0.229 \\
ESD + Ours ($\sigma = 0.15$) & 308 & 0.234 \\
\midrule
Standard ACE & 491 & 0.236 \\
ACE + Ours ($\sigma = 0.00$) & 60 & 0.243 \\
ACE + Ours ($\sigma = 0.05$) & 111 & 0.235 \\
ACE + Ours ($\sigma = 0.10$) & 148 & 0.232 \\
ACE + Ours ($\sigma = 0.15$) & 254 & 0.233 \\
\midrule
Standard EAP & 1142 & 0.236 \\
EAP + Ours ($\sigma = 0.00$) & 638 & 0.234 \\
EAP + Ours ($\sigma = 0.05$) & 677 & 0.235 \\
EAP + Ours ($\sigma = 0.10$) & 734 & 0.231 \\
EAP + Ours ($\sigma = 0.15$) & 883 & 0.237 \\
\midrule
Standard RECELER & 693 & 0.237 \\
RECELER + Ours ($\sigma = 0.00$) & 326 & 0.243 \\
RECELER + Ours ($\sigma = 0.05$) & 374 & 0.241 \\
RECELER + Ours ($\sigma = 0.10$) & 402 & 0.238 \\
RECELER + Ours ($\sigma = 0.15$) & 433 & 0.244 \\
\bottomrule
\end{tabular}
\end{table}

\section{Computational Overhead and Optimization Efficiency}
\label{app:computational_overhead}

Adaptive Noise Sampling introduces a forward-pass scan over $N$ candidate noise vectors but substantially reduces the number of gradient-update steps needed to reach the reported checkpoints. Under standard uniform sampling, both ACE and EAP use 2,000 optimization steps. With our method, ACE uses 500 steps and EAP uses 300 steps, corresponding to reductions of 75\% and 85\%, respectively.

To characterize the computational trade-off, \Cref{tab:scan_time} reports the per-step cost of scanning $N$ candidates, while \Cref{tab:total_time_comparison} reports the total end-to-end runtime. Although our method requires fewer gradient-update steps, the candidate scan increases wall-clock runtime under the current implementation. Because unlearning is a one-time model edit, this modest overhead---hours rather than days---is a practical cost for improved unlearning accuracy, retention, and adversarial robustness. We therefore interpret these results as improved optimization-step efficiency rather than wall-clock acceleration.

\begin{table}[ht]
\centering
\caption{Average time cost (in seconds) per step for the forward-pass exploration scan across $N$ candidates. Because this step relies solely on the frozen guidance classifier and the base diffusion model, this overhead is uniform across all underlying unlearning methods.}
\label{tab:scan_time}
\begin{tabular}{lc}
\toprule
\textbf{Scan Candidates ($N$)} & \textbf{Time (s)} \\
\midrule
4  & 8.9   \\
8  & 17.7  \\
20 & 44.4  \\
40 & 171.8 \\
\bottomrule
\end{tabular}
\end{table}

\begin{table}[ht]
\centering
\caption{Total wall-clock training time for the standard baselines and Adaptive Noise Sampling configured with $N=20$ and $M=4$. The adaptive variants use fewer gradient-update steps but incur additional runtime from candidate scanning.}
\label{tab:total_time_comparison}
\begin{tabular}{lcc}
\toprule
\textbf{Method} & \textbf{Baseline Total Time} & \textbf{Ours} \\
\midrule
ACE & 4h 40m & 7h 20m \\
ESD & 4h 28m & 7h 32m \\
EAP & 3h 22m & 4h 15m \\
RECELER & 4h 32m & 7h 00m \\
\bottomrule
\end{tabular}
\end{table}

\subsection{Hardware and Peak Memory Requirements}
All experiments are conducted on a single NVIDIA A100 GPU with 40\,GB of VRAM. Under our standard configuration ($N=20$, $M=4$), the exploration of the $N$ candidate noise vectors is implemented as a strictly detached forward-pass scan. Consequently, no computation graph for the full diffusion trajectories is retained, and backpropagation is performed only through the selected Top-$M$ samples. The backward-pass memory requirement is therefore bounded by the selected update batch size $M$, rather than by the exploration pool size $N$.

This separation between the sampling strategy and the underlying unlearning optimization prevents peak VRAM usage from increasing drastically. In practice, the peak memory footprint of our method remains comparable to that of the corresponding standard uniform-sampling baseline. \Cref{tab:peak_vram} reports the observed peak VRAM utilization when applying our standard configuration to each baseline method.

\begin{table}[ht]
\centering
\caption{Peak VRAM utilization on a single NVIDIA A100 GPU using Adaptive Noise Sampling with $N=20$ and $M=4$. Because candidate exploration is detached from the unlearning backward pass, memory usage remains comparable to standard uniform-sampling optimization.}
\label{tab:peak_vram}
\begin{tabular}{lc}
\toprule
\textbf{Baseline Method} & \textbf{Peak VRAM (GB)} \\
\midrule
EAP & 24 \\
ACE & 24 \\
ESD & 22 \\
RECELER & 24 \\
\bottomrule
\end{tabular}
\end{table}

\section{Ablation Study: Hyperparameter Sensitivity}
\label{app:ablation}

To systematically evaluate the sensitivity of our Adaptive Noise Sampling framework, we conducted a comprehensive ablation study varying our core hyperparameters: the candidate pool size ($N$), the update batch size ($M$), and the regularization constant ($\gamma$). \Cref{tab:ablation_nm} reports the Failure Count (multi-seed evaluation) across various configurations of ($N, M$) for four baselines: ACE, ESD, EAP, and RECELER.

\begin{table}[ht]
\centering
\caption{Ablation study on the candidate pool ($N$) and update batch ($M$) sizes using NudeNet for guidance and Open-NSFW2 for evaluation. We report the multi-seed Failure Count (lower is better) across four baselines. The standard baselines (top row) represent uniform sampling without our adaptive framework.}
\label{tab:ablation_nm}
\begin{tabular}{cccccc}
\toprule
\textbf{Candidate Pool ($N$)} & \textbf{Update Batch ($M$)} & \textbf{ACE} & \textbf{ESD} & \textbf{EAP} & \textbf{RECELER} \\
\midrule
- (Standard Baseline) & - & 491 & 510 & 1142 & 693 \\
\midrule
4  & 1 & 119 & 265 & 820 & 441 \\
4  & 2 & 111 & 252 & 795 & 425 \\
4  & 4 & 95  & 238 & 762 & 390 \\
\midrule
8  & 1 & 112 & 254 & 805 & 430 \\
8  & 2 & 107 & 245 & 780 & 412 \\
8  & 4 & 89  & 225 & 730 & 375 \\
8  & 8 & 80  & 210 & 705 & 358 \\
\midrule
20 & 1 & 104 & 240 & 770 & 405 \\
20 & 2 & 97  & 228 & 745 & 388 \\
20 & 4 & 60  & 201 & 638 & 326 \\
20 & 8 & 54  & 192 & 602 & 310 \\
\midrule
40 & 1 & 70  & 215 & 690 & 355 \\
40 & 2 & 65  & 208 & 675 & 338 \\
40 & 4 & 55  & 185 & 610 & 305 \\
40 & 8 & 52  & 178 & 595 & 292 \\
\bottomrule
\end{tabular}
\end{table}

\paragraph{Impact of $N$ and $M$:} 
The results in \Cref{tab:ablation_nm} highlight the distinct roles of exploration and exploitation in our framework. Increasing the exploration pool ($N$) allows the framework to discover more elusive, high-activation noise trajectories. Concurrently, increasing the update batch ($M$) provides a stronger, more stable gradient signal for unlearning. While maximizing both parameters naturally yields the lowest failure rates (e.g., $N=40, M=8$ achieves 52 failures on ACE), we observe heavily diminishing returns as computational costs scale. We empirically selected $N = 20$ and $M = 4$ for our primary experiments as it strikes the optimal trade-off: it achieves near-optimal robustness (60 failures on ACE) while strictly bounding the memory overhead of the backward pass to match a standard batch size of 4. 

Importantly, this scaling behavior is highly consistent across all evaluated architectures. Under our selected $N=20, M=4$ configuration, EAP failures are nearly halved from 1142 to 638, and RECELER failures drop dramatically from 693 to 326. Furthermore, the data reveals that even the most computationally lightweight configuration ($N=4, M=1$) vastly outperforms the standard uniform-sampling baselines across the board. For instance, simply updating on the single worst-case noise out of a small pool of 4 candidates reduces ACE failures from 491 to 119. This demonstrates that the core theoretical mechanism of our framework---purposefully guiding the optimization toward high-risk noise regions---is fundamentally effective even without aggressive hyperparameter scaling.

\paragraph{Sensitivity to the Regularization Constant ($\gamma$):}
We additionally evaluated the sensitivity of the framework to the regularization constant $\gamma$. Performance remained consistent across the tested range from $10^{-5}$ to $10^{-3}$. We therefore use $\gamma = 10^{-4}$ for all reported experiments.

\section{Scaling to Larger Architectures: Stable Diffusion 2.1}
\label{app:sd21_scaling}

To verify that our Adaptive Noise Sampling framework scales effectively to larger, more complex generative models, we extend our nudity unlearning experiments to Stable Diffusion 2.1 (SD 2.1). For this evaluation, we integrate our framework with EAP and ESD, both of which provide stable, established support for the SD 2.1 architecture, using NudeNet for adaptive guidance and Open-NSFW2 for evaluation.

As detailed in \Cref{tab:sd21-scaling}, our adaptive framework seamlessly generalizes to this higher-capacity model, delivering substantial improvements over the standard uniform-sampling baselines. When applied to EAP, our method reduces the rigorous 20-seed failure count from 1413 to 985. The gains are even more pronounced when integrated with ESD, where our approach aggressively cuts the 20-seed failure count from 673 down to just 250---a reduction of over 60\%. 

Crucially, these dramatic improvements in unlearning efficacy do not come at the cost of visual fidelity. In fact, our framework slightly improves the CLIP scores for both EAP ($0.233 \to 0.236$) and ESD ($0.221 \to 0.223$). This confirms that dynamically targeting the most relevant regions of the noise manifold successfully protects global generative quality, even as the underlying model capacity and parameter dimensionality increase.

\begin{table}[ht]
\centering
\caption{Nudity unlearning performance scaled to \textbf{Stable Diffusion 2.1} using NudeNet for guidance and Open-NSFW2 for evaluation. We report the Failure Count under standard (1-seed) and rigorous (20-seed) evaluations, alongside the CLIP score for generation quality.}
\label{tab:sd21-scaling}
\begin{tabular}{lccc}
\toprule
\textbf{Method} & \textbf{Failure Count (1-seed) $\downarrow$} & \textbf{Failure Count (20-seed) $\downarrow$} & \textbf{CLIP Score $\uparrow$} \\
\midrule
SD 2.1 (Original) & 130 & 2355 & 0.239 \\
\midrule
Standard EAP & 75 & 1413 & 0.233 \\
EAP + Ours & 43 & 985 & 0.236 \\
\midrule
Standard ESD & 32 & 673 & 0.221 \\
ESD + Ours & 12 & 250 & 0.223 \\
\bottomrule
\end{tabular}
\end{table}

\section{Proof of Concept on a DiT Architecture: Stable Diffusion 3}
\label{app:sd3_dit}

To evaluate whether Adaptive Noise Sampling generalizes beyond U-Net-based diffusion models, we conduct a proof-of-concept experiment on Stable Diffusion 3 (SD3), which uses a Diffusion Transformer (DiT) architecture. We use Direct Unlearning Optimization (DUO)~\cite{NEURIPS2024_92f43b1d}, a DiT-compatible unlearning method, as the baseline and apply our sampling strategy without modifying the underlying DUO objective. We use NudeNet for adaptive guidance and Open-NSFW2 for evaluation under both the standard single-seed protocol and the more rigorous 20-seed protocol.

As shown in \Cref{tab:sd3_dit}, Adaptive Noise Sampling improves the robustness of DUO on SD3. It reduces the single-seed failure count from 60 to 46, a $23.3\%$ reduction, and the 20-seed failure count from 1245 to 938, a $24.7\%$ reduction. At the same time, the CLIP score increases slightly from $0.2385$ to $0.2399$, indicating that the improved erasure does not compromise text--image alignment. These results provide initial evidence that our framework transfers to DiT architectures without architecture-specific modifications and reduces the probabilistic tail risk associated with unfavorable noise initializations.

\begin{table}[ht]
\centering
\caption{Proof-of-concept nudity unlearning results on \textbf{Stable Diffusion 3} using NudeNet for guidance and Open-NSFW2 for evaluation. We compare standard DUO~\cite{NEURIPS2024_92f43b1d} with DUO augmented by our Adaptive Noise Sampling strategy. Lower failure counts and higher CLIP scores are better.}
\label{tab:sd3_dit}
\begin{tabular}{lccc}
\toprule
\textbf{SD3 Unlearning Method} & \textbf{Failures (1 seed) $\downarrow$} & \textbf{Failures (20 seeds) $\downarrow$} & \textbf{CLIP Score $\uparrow$} \\
\midrule
Standard DUO Baseline & 60 & 1245 & 0.2385 \\
DUO + Ours & 46 & 938 & 0.2399 \\
\bottomrule
\end{tabular}
\end{table}

\section{Unlearning on FLUX.1 [dev]}
\label{app:flux_unlearning}

We further evaluate Adaptive Noise Sampling on FLUX.1 [dev], using the unlearning method from EraseAnything~\cite{gao2025eraseanything} as the baseline. We compare the original model, standard EraseAnything, and EraseAnything augmented with our adaptive sampling strategy on nudity unlearning, using NudeNet for adaptive guidance and Open-NSFW2 for evaluation. We report failure counts for the single-seed and 20-seed evaluations, comprising 200 and 4,000 generated images, respectively, together with the mean Open-NSFW2 score for the 20-seed evaluation and the CLIP score.

\begin{table}[ht]
\centering
\caption{Nudity unlearning results on \textbf{FLUX.1 [dev]} using NudeNet for guidance and Open-NSFW2 for evaluation. Failure counts are reported out of 200 images (1 seed) and 4,000 images (20 seeds). Values are point estimates; run-to-run standard deviations are not reported. A dash denotes an unavailable mean score. Bold indicates the better result between standard EraseAnything and EraseAnything + Ours.}
\label{tab:flux_unlearning}
\adjustbox{max width=\textwidth}{
\begin{tabular}{lcccc}
\toprule
Method & Failures (1 seed) $\downarrow$ & Failures (20 seeds) $\downarrow$ & Mean score (20 seeds) $\downarrow$ & CLIP $\uparrow$ \\
\midrule
FLUX.1 [dev] (Original) & 75/200 & 1554/4000 & --- & 0.235414 \\
EraseAnything & 48/200 & 1002/4000 & 0.257866 & 0.222557 \\
EraseAnything + Ours & \textbf{35/200} & \textbf{703/4000} & \textbf{0.214151} & \textbf{0.227629} \\
\bottomrule
\end{tabular}
}
\end{table}

As shown in \Cref{tab:flux_unlearning}, our adaptive strategy reduces the single-seed failure count from 48 to 35 (a $27.1\%$ reduction) and the 20-seed failure count from 1002 to 703 (a $29.8\%$ reduction) relative to standard EraseAnything. The mean nudity score also decreases from $0.257866$ to $0.214151$, a $17.0\%$ reduction. At the same time, the CLIP score increases from $0.222557$ to $0.227629$ ($+2.3\%$), although it remains below the original model's score of $0.235414$. These results support the applicability of Adaptive Noise Sampling to FLUX.1 [dev], improving erasure while partially recovering text--image alignment relative to the unlearning baseline. Residual failures remain, so these improvements do not imply complete concept removal.

\section{Adversarial Evaluation}
\label{sec:adversarial_evaluation}

We further evaluate whether the robustness gains from Adaptive Noise Sampling persist under adversarial attacks designed to recover erased concepts. For each underlying unlearning method, we attack both its standard checkpoint (\emph{Base}) and the corresponding checkpoint trained with our adaptive sampling strategy (\emph{Ours}). This paired comparison isolates the contribution of Adaptive Noise Sampling while holding the underlying unlearning objective fixed. We consider the black-box Ring-A-Bell attack~\cite{tsai2024ring} and the white-box UnlearnDiffAtk (UDA) attack~\cite{zhang2024generate}. We report attack success rate (ASR), which provides a normalized and directly comparable measure of adversarial robustness across attacks and evaluation settings.

\Cref{tab:adversarial_attacks} summarizes the results for both attacks across ACE, RECELER, ESD, and EAP. Under Ring-A-Bell, Adaptive Noise Sampling reduces the ASR from 0.20\% to 0.00\% for ACE, 1.38\% to 0.43\% for RECELER, 45.68\% to 35.55\% for ESD, and 53.13\% to 32.38\% for EAP. In the corresponding single-seed evaluation, ESD failures decrease from 92/200 to 80/200 and EAP failures from 107/200 to 76/200. Under UnlearnDiffAtk (UDA), the ASR decreases from 2.81\% to 2.42\% for ACE, 19.14\% to 11.88\% for RECELER, 76.35\% to 23.28\% for ESD, and 21.74\% to 15.58\% for EAP.

\begin{table}[ht]
\centering
\caption{Adversarial nudity-unlearning evaluation under the black-box Ring-A-Bell attack~\cite{tsai2024ring} and the white-box UnlearnDiffAtk (UDA) attack~\cite{zhang2024generate}. We compare each standard unlearning checkpoint (Base) with the checkpoint trained using Adaptive Noise Sampling (Ours). We report attack success rate (ASR; lower is better). Ring-A-Bell ASRs are computed from the 20-seed evaluation over 4,000 generations.}
\label{tab:adversarial_attacks}
\adjustbox{max width=\columnwidth}{
\renewcommand{\arraystretch}{1.08}
\setlength{\tabcolsep}{10pt}
\begin{tabular}{llc}
\toprule
Attack & Method & ASR (\%) $\downarrow$ \\
\midrule
\multirow{8}{*}{\shortstack[l]{\textbf{Black-box}\\Ring-A-Bell~\cite{tsai2024ring}}}
& ACE Base                 & 0.20 \\
& \textbf{ACE + Ours}     & \textbf{0.00} \\
\addlinespace[3pt]
& RECELER Base             & 1.38 \\
& \textbf{RECELER + Ours} & \textbf{0.43} \\
\addlinespace[3pt]
& ESD Base                 & 45.68 \\
& \textbf{ESD + Ours}     & \textbf{35.55} \\
\addlinespace[3pt]
& EAP Base                 & 53.13 \\
& \textbf{EAP + Ours}     & \textbf{32.38} \\
\midrule
\multirow{8}{*}{\shortstack[l]{\textbf{White-box}\\UnlearnDiffAtk (UDA)~\cite{zhang2024generate}}}
& ACE Base                 & 2.81 \\
& \textbf{ACE + Ours}     & \textbf{2.42} \\
\addlinespace[3pt]
& RECELER Base             & 19.14 \\
& \textbf{RECELER + Ours} & \textbf{11.88} \\
\addlinespace[3pt]
& ESD Base                 & 76.35 \\
& \textbf{ESD + Ours}     & \textbf{23.28} \\
\addlinespace[3pt]
& EAP Base                 & 21.74 \\
& \textbf{EAP + Ours}     & \textbf{15.58} \\
\bottomrule
\end{tabular}
}
\end{table}

\section{Additional NudeNet Training and Evaluation Results}\label{sec:nudenet}

This section intentionally reports a same-classifier diagnostic in which NudeNet~\cite{nudenet} serves as both the training-time guidance classifier and the evaluation classifier. Unlike the main experiment, which evaluates NudeNet-guided checkpoints independently with Open-NSFW2, this analysis measures performance directly under the detector used to guide unlearning. We follow the same protocol as in the main paper, reporting failure counts under a single predefined seed and under 20 randomly sampled noise seeds per prompt.

\begin{table}[H]
\centering
\caption{Nudity unlearning results evaluated using NudeNet~\cite{nudenet}. For our adaptive variants, NudeNet is used for both training guidance and evaluation. \textit{Failure count} (lower is better) denotes the number of images classified as nudity out of 200 prompts (single predefined seed) or 4,000 images (20 random seeds per prompt).}
\label{tab:nudity_nudenet}
\adjustbox{max width=\columnwidth}{
\begin{tabular}{lcc}
\toprule
Method &
Failure (1 seed) $\downarrow$ &
Failure (20 seeds) $\downarrow$ \\
\midrule
SD 1.4 (Original) & 137 & 2700 \\
UCE              & 35  & 686  \\
ESD              & 9   & 212  \\
ACE              & 11  & 227  \\
EAP              & 21  & 395  \\
RECELER          & 7   & 251  \\
\midrule
\textbf{Ours (ESD; NudeNet guidance)} &
\textbf{3}\ (\good{-66.7\%}) &
\textbf{93}\ (\good{-56.1\%}) \\
\textbf{Ours (ACE; NudeNet guidance)} &
\textbf{2}\ (\good{-81.8\%}) &
\textbf{18}\ (\good{-92.1\%}) \\
\textbf{Ours (EAP; NudeNet guidance)} &
\textbf{14}\ (\good{-33.3\%}) &
\textbf{276}\ (\good{-30.1\%}) \\
\textbf{Ours (RECELER; NudeNet guidance)} &
\textbf{2}\ (\good{-71.4\%}) &
\textbf{102}\ (\good{-59.4\%}) \\
\bottomrule
\end{tabular}
}
\end{table}

\section{Human evaluation.}\label{sec:human_eval_appendix}

To further validate our findings beyond automatic classifiers, we conduct a human evaluation on nudity unlearning. We randomly select 100 prompts from the nudity subset and generate images using 20 random seeds per prompt, resulting in 2000 images per method. Five annotators (a mix of undergraduate and graduate students) were recruited for the study. Prior to annotation, participants were explicitly informed that the images may contain nudity and provided consent to proceed.

Each image was independently annotated with a binary label indicating whether nudity is present. \Cref{tab:human_eval} reports the total number of images labeled as nude (failure count) and the corresponding failure ratio. Consistent with the automatic evaluation, our method substantially reduces the presence of nudity compared to all baselines. In particular, when applied on top of ACE and ESD, our approach achieves the lowest failure rates, indicating more reliable suppression of the target concept across diverse noise initializations. These results confirm that the improvements observed with automated detectors translate to human perception, strengthening the practical relevance of our unlearning framework.

\begin{table}[ht]
\centering
\caption{Human evaluation results for nudity unlearning. Human annotators provided binary labels (nude / non-nude) for 100 prompts evaluated over 20 random seeds, yielding 2,000 images per method and 20,000 images across the ten evaluated methods. We report the number of images classified as nude for each method (\textit{Failure count}, lower is better) and the corresponding failure rate (\textit{Failure ratio}, lower is better).}
\label{tab:human_eval}
\adjustbox{max width=\columnwidth}{
\begin{tabular}{lcc}
\toprule
Method & Failure count $\downarrow$ & Failure ratio $\downarrow$ \\
\midrule
SD 1.4 (Original) & 1800 & 0.90 \\
UCE               & 993  & 0.497 \\
ESD               & 480  & 0.240 \\
ESD+ (Ours)       & 55  & 0.0275 \\
ACE               & 491  & 0.246 \\
ACE+ (Ours)       & 24   & 0.012 \\
EAP               & 739  & 0.370 \\
EAP+ (Ours)       & 372  & 0.186 \\
RECELER           & 660  & 0.330 \\
RECELER+ (Ours)   & 435  & 0.218 \\
\bottomrule
\end{tabular}
}
\end{table}

\section{Additional Qualitative Results}
\label{sec:additional_qualitative}

We provide additional qualitative results to complement the quantitative evaluations in the main paper, highlighting the behavior of our noise-aware unlearning strategy under controlled and stochastic settings. In particular, we examine both (i) fixed prompt–seed conditions and (ii) variations across different unlearning methods and noise initializations.

\begin{figure*}[ht]
    \centering
    \includegraphics[width=\textwidth]{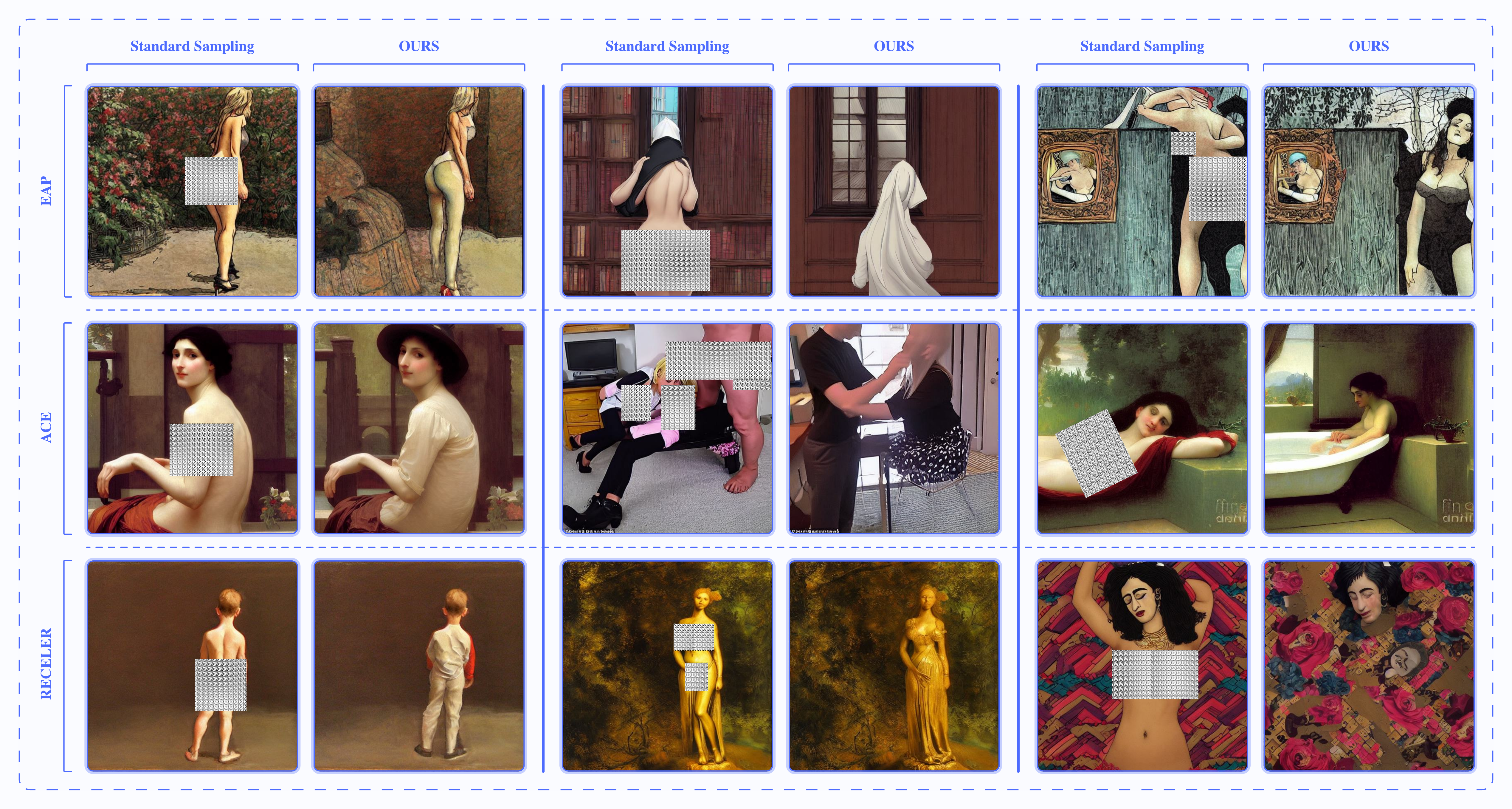}
    \caption{Qualitative comparison under the same prompt and identical noise seed. Our method generates images that are visually consistent with the baseline output while effectively removing the target concept, demonstrating precise and controlled unlearning without altering overall image structure.}
    \label{fig:qualitative_more}
\end{figure*}

\Cref{fig:qualitative_more} presents results generated using the \emph{same prompt and the same random seed} across methods. Under identical noise initialization, our approach produces images that remain visually consistent with the baseline generation—preserving scene layout, structure, and semantic content—while successfully removing the forbidden concept. This demonstrates that our method performs targeted unlearning without introducing unnecessary deviations or artifacts when the stochastic factors are held constant.

\begin{figure*}[ht]
    \centering
    \includegraphics[width=0.98\columnwidth]{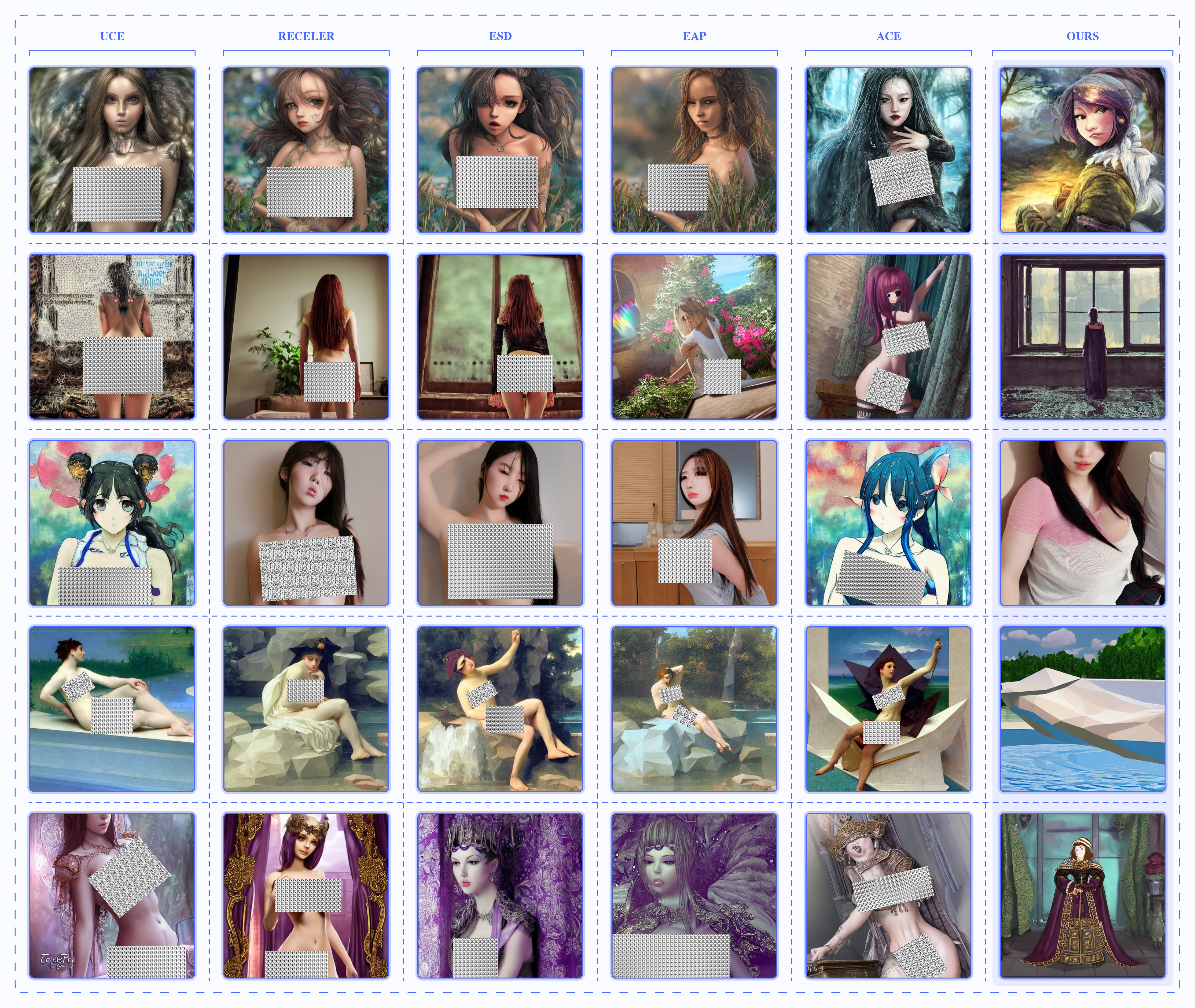}
    \caption{Qualitative comparison across different unlearning methods under the same prompt. Our approach more consistently suppresses the target concept across stochastic noise initializations, while preserving semantic content and visual quality compared to baseline methods.}
    \label{fig:qualitative_more2}
\end{figure*}

\Cref{fig:qualitative_more2} compares different unlearning methods under the \emph{same prompt}. While baseline approaches often exhibit residual appearances of the target concept or noticeable degradation in image quality, our method more reliably suppresses the concept across stochastic samples, maintaining visual plausibility and prompt faithfulness. Together, these examples illustrate that explicitly accounting for noise-space structure leads to more robust and stable unlearning behavior under both deterministic and stochastic generation settings.

\end{document}